\documentclass{article}
\usepackage{times}
\usepackage{soul}
\usepackage{url}
\usepackage[hidelinks]{hyperref}
\usepackage[utf8]{inputenc}
\usepackage[small]{caption}
\usepackage{graphicx}
\usepackage{amsmath}
\usepackage{amsthm}
\usepackage{booktabs}
\usepackage{algorithm}
\usepackage{algorithmic}
\usepackage[algo2e]{algorithm2e}
\usepackage{verbatim}
\usepackage{graphicx}
\usepackage{amsmath}
\usepackage{bm}
\usepackage{booktabs}
\usepackage{subcaption}
\usepackage{textcomp}
\usepackage{multirow}
\usepackage{color}

\usepackage{wrapfig}

\usepackage{color,url}
\usepackage{multicol}
\usepackage{multirow}

\usepackage{amsmath,epsfig,bm, amssymb,graphicx,algorithm,algorithmic,color}

\newcommand{\mathbbR}{\mathbb{R}}

\newcommand{\boldA}{{\boldsymbol{A}}}

\newcommand{\boldM}{{\boldsymbol{M}}}

\newcommand{\boldW}{{\boldsymbol{W}}}
\newcommand{\boldX}{{\boldsymbol{X}}}

\newcommand{\bolde}{{\boldsymbol{e}}}

\newcommand{\boldg}{{\boldsymbol{g}}}
\newcommand{\boldh}{{\boldsymbol{h}}}

\newcommand{\boldu}{{\boldsymbol{u}}}

\newcommand{\boldx}{{\boldsymbol{x}}}
\newcommand{\boldy}{{\boldsymbol{y}}}

\newcommand{\bolddelta}{{\boldsymbol{\delta}}}

\newcommand{\boldtheta}{{\boldsymbol{\theta}}}

\newcommand{\boldmu}{{\boldsymbol{\mu}}}

\newcommand{\boldphi}{{\boldsymbol{\phi}}}

\newcommand{\boldomega}{{\boldsymbol{\omega}}}

\newcommand{\boldDelta}{{\boldsymbol{\Delta}}}
\newcommand{\boldTheta}{{\boldsymbol{\Theta}}}

\newcommand{\boldvarphi}{{\boldsymbol{\varphi}}}

\newcommand{\calS}{{\mathcal{S}}}

\usepackage{color}

\advance\oddsidemargin-1in
\title{FsNet: Feature Selection Network on \\ High-dimensional Biological Data}

\author{
Dinesh Singh$^1$, H\'{e}ctor Climente-Gonz\'{a}lez$^1$, Mathis Petrovich$^2$,\\ Eiryo Kawakami$^{3,4}$,
Makoto~Yamada$^{1,5}$\\
$^1$RIKEN AIP, $^2$\'{E}cole des Ponts ParisTech, \\
$^3$RIKEN Medical Sciences Innovation Hub Program, $^4$Chiba University\\
$^5$Kyoto University,
}

\begin{document}

\maketitle

\begin{abstract}
    Biological data including gene expression data are generally high-dimensional and require efficient, generalizable, and scalable machine-learning methods to discover their complex nonlinear patterns. The recent advances in machine learning can be attributed to deep neural networks (DNNs), which excel in various tasks in terms of computer vision and natural language processing. However, standard DNNs are not appropriate for high-dimensional datasets generated in biology because they have many parameters, which in turn require many samples. In this paper, we propose a DNN-based, nonlinear feature selection method, called the feature selection network (FsNet), for high-dimensional and small number of sample data. Specifically, FsNet comprises a selection layer that selects features and a reconstruction layer that stabilizes the training. Because a large number of parameters in the selection and reconstruction layers can easily result in overfitting under a limited number of samples, we use two tiny networks to predict the large, virtual weight matrices of the selection and reconstruction layers. Experimental results on several real-world, high-dimensional biological datasets demonstrate the efficacy of the proposed method. The Python code is available at \url{https://github.com/dineshsinghindian/fsnet}
\end{abstract}

\section{Introduction}
    The recent advancements in measuring devices for life sciences have resulted in the generation of large biological datasets, which are extremely important for many medical and biological applications, including disease diagnosis, biomarker discovery, drug development, and forensics~\cite{Li2014gpb}. Generally, such datasets are substantially high-dimensional (i.e., many features with small number of samples) and contain complex nonlinear patterns. Machine learning methods, including genome-wide association studies ($d > 10^5$, $n < 10^4$) and gene selection ($d > 10^4$, $n < 10^3$)~\cite{Vivien2013nature}, have been successfully applied to discover the complex patterns hidden in high-dimensional biological and medical data. However, most nonlinear models in particular deep neural networks (DNN) are difficult to train under these conditions because of the significantly high number of parameters. Hence, the following questions naturally arise: 1) are all the features necessary for building effective prediction models? and 2) what modifications are required in the existing machine-learning methods to efficiently process such high-dimensional data?

    The answer to the first question is to select the most relevant features, thereby requiring an appropriate feature selection method~\cite{Ye2019ijcai,Ming2019ijcai,Liao2019ijcai}. This problem, called feature selection, consists of identifying a smaller subset (i.e., smaller than the original dataset) that contains relevant features such that the subset retains the predictive capability of the data/model while eliminating the redundant or irrelevant features~\cite{yamada2014high,Yamada2018tkde,10.1093/bioinformatics/btz333}. Most state-of-the-art feature selection methods are based on either sparse-learning methods, including Lasso \cite{Tibshirani1996jrss}, or kernel methods~\cite{Masaeli2010icml,Yamada2018aistats,yamada2014high}. These \emph{shallow} approaches satisfactorily work in practice for biological data. However, sparse-learning models including Lasso are in general linear and hence cannot capture high-dimensional biological data. Kernel-based methods can handle the nonlinearlity, but it heavily depends on the choice of the kernel function. Thus, more flexibile approaches that can train an arbitrary nonlinear transformation of features are desired.
    
    An approach to learning such a nonlinear transformation could be based on deep autoencoders~\cite{Vincent2010jmlr}. However, deep autoencoders are useful for computer-vision and natural language processing tasks, wherein a large number of training samples are available. In contrast, for high-dimensional biological data, the curse of dimensionality prevents us from training such deep models without overfitting. Moreover, these models focus on building useful features rather than selecting features from data. The training of autoencoders for feature selection results in the discrete combinatorial optimization problem, which is difficult to train in an end-to-end manner.

    To train neural networks on high-dimensional data without resulting in overfitting, several approaches were proposed. Widely used ones are based on random projection and its variants~\cite{Dahl2013icassp,Wojcik2019paa}. However, their performances significantly depend on the random projection matrix, and their usability is limited to dimensionality reduction only. Therefore, they cannot be applied for feature selection. Another deep learning-based approach employs a concrete autoencoder (CAE)~\cite{Balin2019icml}, which uses concrete random variables~\cite{Maddison2017iclr} to select features without supervision. Although CAE is an unsupervised model with poor performance, it can be extended to incorporate a supervised-learning setup. However, we observed that this simple extension is not efficient because the large number of parameters in the first layer of CAE can easily result in overfitting under a limited number of samples. 

    To address these issues, we propose a non-linear feature selection network, called FsNet, for high-dimensional biological data. FsNet comprises a selection layer that uses concrete random variables~\cite{Maddison2017iclr}, which are the continuous variants of a one-hot vector, and a reconstruction layer that stabilizes the training process. The concrete random variable allows the conversion of the discrete optimization problem into a continuous one, enabling the backpropagation of gradients using the reparameterization trick. During the training period, FsNet selects a few features using its selection layer while maximizing the classification accuracy and minimizing the reconstruction error. However, owing to the large number of parameters in the selection and reconstruction layers, overfitting can easily occur under a limited number of samples. Therefore, to avoid overfitting, we propose using two tiny networks to predict the large, virtual weight matrices of the selection and reconstruction layers. Consequently, the size of the model is significantly reduced and the network can scale high-dimensional datasets on a resource-limited device/machine. Through experiments on various real-world datasets, we show that the proposed FsNet significantly outperforms CAE and the supervised counterpart thereof.

    \noindent {\bf Contributions:} Our contributions through this paper are as follows.
    \begin{itemize}
        \item We propose FsNet, an end-to-end trainable neural network based nonlinear feature selection, for high-dimensional data with small number of samples. 
        \item  FsNet compares favorably with the state-of-the-art nonlinear feature selection methods  for high-dimensional data with small number of samples. 
        \item The model size of FsNet is one to two orders magnitude smaller than that of a standard DNN model, including CAE \cite{Balin2019icml}.
    \end{itemize}
    
\section{Related Work}
    Here, we discuss the existing shallow/deep feature selection methods, along with their drawbacks. 
    
    \noindent \textbf{Shallow, nonlinear feature selection:} \emph{Maximum relevance} is a simple but effective criterion of nonlinear feature selection \cite{guyon2003introduction}. It uses mutual information and the Hilbert-Schmidt Independence Criterion (HSIC) to select the features associated with the outcome \cite{Peng2005pami,song2007supervised}. It is also called sure independence screening in the statistics community \cite{fan2008sure,Balasubramanian2013aistats}. However, because it tends to select redundant features, minimum redundancy maximum relevance (mRMR) feature selection was proposed~\cite{Peng2005pami}. Notably, mRMR finds the subset of independent features that are maximally associated with the outcome by using mutual information between features and between each feature and the outcome. Recently, a kernel-based, convex variant of mRMR was proposed, called HSIC Lasso \cite{yamada2014high,Yamada2018tkde,10.1093/bioinformatics/btz333}. They effectively perform nonlinear feature selection on high-dimensional data, producing simple models with parameters that can be easily estimated. However, their performances are limited by the simplicity of the models and depends on the choice of kernels.

    \noindent \textbf{DNNs for feature selection:}
    DNNs are nonlinear, complex models that can address the aforementioned problems associated with kernel-based methods. They can be used for feature selection by adding a regularization term to the loss function, or by measuring the effect of an input feature on the target variable~\cite{Verikas2002prl}. Elaborately, an extra feature scoring layer is added to perform element-wise multiplication on the features and score, and then they are entered as inputs into the rest of the network~\cite{Wang2014nips,Lu2018nips}. However, DNNs do not select features during the training period, thereby resulting in a performance reduction after feature selection. Moreover, it is generally difficult to obtain a sparse solution using a stochastic gradient. CAE~\cite{Balin2019icml} addresses this problem by training an autoencoder that contains a feature selection layer with a concrete variable, which is a continuous relaxation of a one-hot vector. Recently, another end-to-end, supervised, feature selection method based on stochastic gates (STGs) was proposed \cite{yamada2020feature}. It uses a continuously relaxed Bernoulli variable and performs better than the existing feature selection methods. However, these methods need to train a large number of parameters in the first layer, resulting in overfitting to the training data. Therefore, these approaches may not be appropriate for DNN models with high-dimensional data and a limited number of samples.

    \noindent \textbf{Training DNNs on high-dimensional data:} 
    The existing DNN-based methods can easily overfit to the high-dimensional biological data, as they suffer from the \textit{curse-of-dimensionality} irrespective of \textit{regularization constraints}. The biggest drawback of DNNs is that they need to have a large number of parameters in the first layers of the decoder and encoder. HashedNets~\cite{Chen2015icml} addressed this issue by exploiting the inherent redundancy in weights to group them into relatively fewer hash buckets and shared them with all its connections. However, the hash function groups the weights on the basis of their initial values instead of opting for a dynamic grouping, thereby reducing the options to arbitrarily learn weights. Diet Networks~\cite{Romero2017iclr} used tiny networks to predict weight matrices. However, they are limited to the multilayer perceptron only for classification and not for feature selection. A DNN model, referred to as deep neural pursuit (DNP)~\cite{Liu2017ijcai}, selects features from high-dimensional data with a small number of samples. It is based on changes in the average gradients with multiple dropouts by an individual feature. However, \cite{Liu2017ijcai}  reported that the performance of DNP significantly depends on the number of layers. 

    These issues render the existing approaches inefficient for processing biological data, thereby raising the need to develop a method for efficiently extracting features from biological data.

\section{Problem Formulation}
    Let $\boldX=(\boldx_1,\cdots,\boldx_n)^\top = (\boldu_1, \cdots, \boldu_d) \in \mathbb{R}^{n\times d}$ be the given data matrix, where $\boldx\in\mathbbR^{d}$ represents the sample vector with $d$ number of features and $\boldu\in \mathbbR^n$ the feature vector with $n$ number of samples. Let $\boldy=(y_1,\cdots,y_n)^\top\in\mathbb{R}^n$ be the target vector such that $y_i\in\mathcal{Y}$ represents the output for $\boldx_i$, where $\mathcal{Y}$ denotes the domain of the output vector $\boldy$, which is continuous for regression problems and categorical for classification problems. In this paper, we assume that the number of samples is significantly fewer than that of the dimensions (i.e., $n \ll d$).
    
    The final goal of this paper is to train a neural-network classifier $f(\cdot):\mathbb{R}^d\rightarrow\mathcal{Y}$, which simultaneously identifies a subset $\mathcal{S} \subseteq \mathcal{F} = \{1,2\cdots d\}$ of features of a specified size $|\mathcal{S}|=K\ll d$, where the subset can reproduce the remaining $\mathcal{F}\backslash\mathcal{S}$ features with minimal loss.

    \begin{figure*}[!t]
        \centering
        \includegraphics[width=1\linewidth,keepaspectratio]{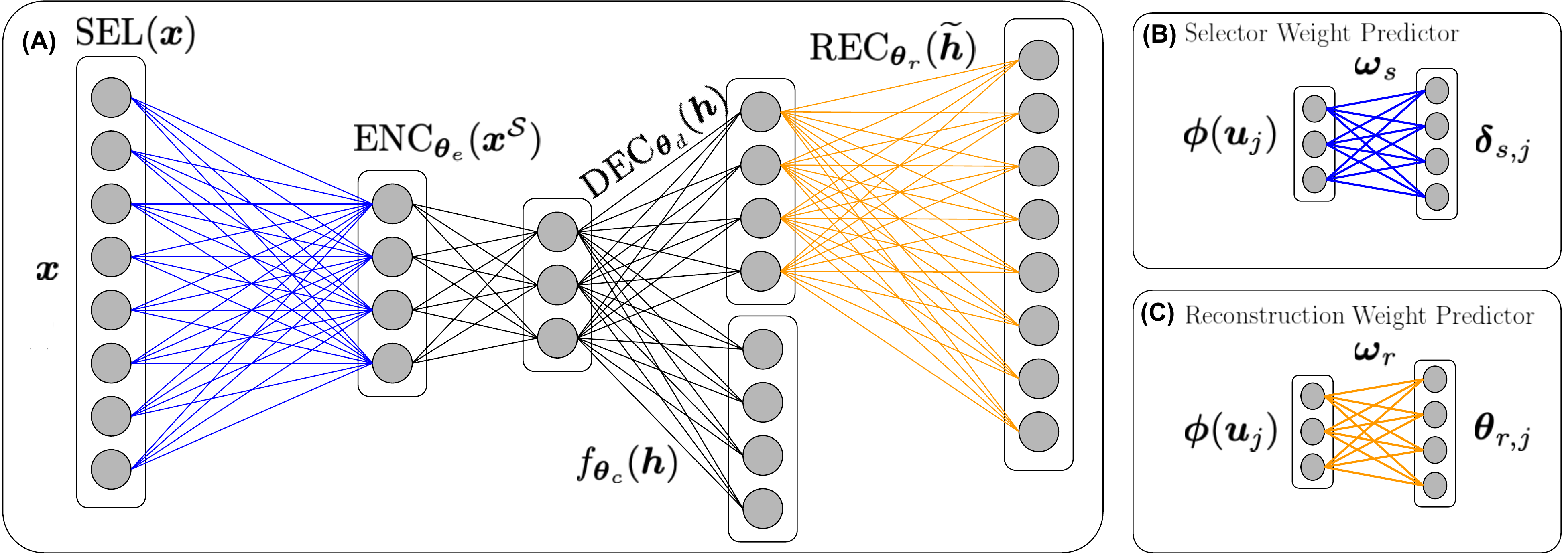}
        \vspace{-.2in}
        \caption{(A) Architecture of FsNet. (B) and (C) are the weight-predictor networks for the selection and reconstruction layers, respectively.
        }
        \label{fig:architecture}
    \end{figure*}
\section{Proposed Method: FsNet}
 
    We here present the architecture and training of the proposed FsNet model for selecting nonlinear features from high-dimensional data.

    \subsection{FsNet Model}
    We aim to build an end-to-end, trainable, compact, feature selection model. Hence, we employ a concrete random variable \cite{Maddison2017iclr} to select features, and we also use the weight-predictor models used in Diet Networks to reduce the model size \cite{Romero2017iclr}. We build FsNet, a simple but effective model (see Figure~\ref{fig:architecture}). As shown in Figure~\ref{fig:architecture}(A), although the selection and reconstruction layers have many connections, they are virtual layers whose weights are predicted from significantly small networks, as shown in Figures~\ref{fig:architecture}(B) \& (C), respectively. The weight-predictor networks (B) and (C) are trained on the feature embeddings. 
    
    The optimization problem of FsNet is given by
    \begin{align}\label{eq:objdis}
        \min_{\boldTheta}&\sum_{i = 1}^n \textnormal{Loss}(y_i,f_{\boldtheta_c}(\textnormal{ENC}_{\boldtheta_e}(\boldx_i^\calS)))+\lambda\sum_{i = 1}^n\|\boldx_i-\textnormal{REC}_{\boldtheta_r}(\textnormal{DEC}_{\boldtheta_d}(\textnormal{ENC}_{\boldtheta_e}(\boldx_i^\calS)))\|_2^2,
    \end{align}
    where $\textnormal{Loss}(y,f_{\boldtheta_c})$ denotes the categorical cross-entropy loss (between $y$ and $f_{\boldtheta_c}$), $\|\cdot\|_2$ the $\ell_2$ norm, $\lambda \geq 0$ the regularization parameter for the reconstruction loss, $\boldTheta$ all the parameters in the model, $\textnormal{SEL}(\cdot)$ the selection layer, $\boldx_i^\calS = \textnormal{SEL}(\boldx_i)$, $\textnormal{ENC}(\cdot)$ the encoder network, $\textnormal{DEC}(\cdot)$ the decoder network, and $\textnormal{REC}(\cdot)$ the reconstruction layer. The \textit{pseudocode} for the training of FsNet is provided in Algorithm~\ref{alg:FsNet} in the supplementary material. 
    
    \noindent {\bf Selection Layer (Train):} We first describe the selection layer, which is used to select important features in an end-to-end manner. The feature selection problem is generally a combinatorial problem, but it is difficult to train in an end-to-end manner because it breaks the propagation of the gradients.  To overcome this obstacle, a concrete random variable~\cite{Maddison2017iclr}, which is a continuous relaxation of a discrete one-hot vector, can be used for the training, as it computes the gradients using the reparameterization trick. Specifically, selecting the $k$-th feature of the input $\boldx$ can be expressed as $\boldx^{(k)} = \bolde_k^\top \boldx$, where $\bolde_k \in \mathbbR^d$ denotes the one-hot vector whose $k$-th feature is 1 and 0 otherwise. The concrete variables for the $k^{th}$ neuron in the selection layer are defined as follows:
    \begin{equation}\label{eq:crv}
        \boldmu^{(k)}=\frac{\exp{((\log{\bolddelta_{s}^{(k)}}+\boldg)/\tau)}}{\sum_{j=1}^d\exp{((\log{\delta_{sj}^{(k)}}+g_j)/\tau)}}, k = 1, 2, \ldots, K,
    \end{equation}
    where $\boldg\in\mathbb{R}^d$ is drawn from the Gumbel distribution. Additionally, $\tau$ denotes the temperature that controls the extent of the relaxation, $K$ the number of selected features, and $\boldDelta_{s}=(\bolddelta_{s,1}, \ldots, \bolddelta_{s,d})= (\bolddelta_{s}^{(1)}, \ldots, \bolddelta_{s}^{(K)})^\top \in \mathbbR^{K \times d}$, $\bolddelta_s^{(k)} \in \mathbbR^K_{>0}$ is the model parameter for concrete variables. Notably, $\boldmu^{(k)}$ becomes a one-hot vector when $\tau \rightarrow 0$.  
    
    Using the concrete variables $\boldM = (\boldmu^{(1)}, \boldmu^{(2)}, \ldots, \boldmu^{(K)})^\top$, the feature selection process can be simply written by using matrix multiplications as follows:
    \begin{align*}
        \textnormal{SEL}(\boldx) = \boldM \boldx.
    \end{align*}
    
    Because the feature selection process can be written by using matrix multiplications, it can be trained in an end-to-end manner. However, the number of parameters in the selection layer is $O(dK)$; it depends on the size of the input layer $d$ and the number of neurons in the selection layer $K$. Thus, for high-dimensional data, the number of model parameters can be high, resulting in overfitting under a limited number of samples $n$. We address both the issues by using a tiny weight-predictor network $\boldvarphi_{\boldomega_s}(\cdot):\mathbbR^b\rightarrow\mathbbR^K_{>0}$ to predict the weights $\bolddelta_{s,j} = \boldvarphi_{\boldomega_s}(\boldphi(\boldu_j))$ (see Figure~\ref{fig:architecture}(B)), where  $\boldphi(\boldu_j)\in\mathbbR^b$ is the embedding representation of feature $j$ and $b\leq n$ the size of the embedding representation. Specifically, the feature embedding $\boldphi(\boldu_j)$ for the $j^{th}$ feature vector used for training the weight-predictor networks is defined as    $\boldphi(\boldu_j)=\boldsymbol{\rho}_j\odot \boldsymbol{\nu}_j$, where $\odot$ denotes elementwise multiplication, whereas $\boldsymbol{\rho}_j$ and $\boldsymbol{\nu}_j$ denote the frequencies and means of the histogram bins of feature $\boldu_j$, respectively. In this paper, we use $\bolddelta_{s,j} = \textnormal{softmax}(\boldW_{\boldomega_s} \boldphi(\boldu_j))$, where $\boldW_{\boldomega_s} \in \mathbbR^{K \times b}$ is the model parameter for the tiny network. Over epochs, $\boldmu^{(k)}$ will converge to a one-hot vector. {Notably, the model parameter $\boldDelta \in \mathbbR^{K \times d}$ depends on the input dimension $d$. However because the model size of the weight-predictor network depends on $b \ll d$, we can significantly reduce the network model size using the predictor network. Moreover, the tiny weight-predictor network can also be trained in an end-to-end manner.}

    \begin{wrapfigure}[12]{r}[5mm]{85mm}
    \vspace{-.2in}
    \begin{minipage}{0.45\textwidth}
      \begin{algorithm}[H]
        \caption{Unique argmax function $\textnormal{uargmax}$}
        \label{alg:uargmax}
        \textbf{Input}: matrix $\boldA\in\mathbb{R}_+^{d\times K}$, with $d$ rows and $K$ cols\\
        \textbf{Output}: selected indices $\mathcal{S}$ %
        \begin{algorithmic}[1] 
            \STATE $\mathcal{S}\leftarrow\{\}$
            \FOR{$i=0 - K$}
            \STATE $(x,y)\leftarrow\text{ index of max value in }\boldA$\\
            \STATE $\mathcal{S}\leftarrow \mathcal{S}\cup x$\\
            \STATE $A.row(x)\leftarrow \mathbf{0}$\\
            \STATE $A.col(y)\leftarrow \mathbf{0}$\\
            \ENDFOR
        \end{algorithmic}
    \end{algorithm}
    \end{minipage}
  \end{wrapfigure}
    
    \noindent {\bf Selection Layer (Inference):} {For inference, we can replace the concrete variables with a set of feature indices. Consequently, the inference becomes faster than before, as we need not compute tiny networks.
    However, if we simply use the argmax function, it tends to select redundant features, and thus the prediction performance can be degraded. Therefore, we propose the \textit{unique\_argmax} function to select non-redundant features and then use the non-redundant feature set for inference}.  The $K$ best and unique features are selected from the estimated $\boldM$ as $\mathcal{S} = \ \textnormal{uargmax}(\boldM^\top)$. Subsequently, for inference, we use $\boldx^\calS \in \mathbbR^K$ as an input of the encoder network. Although this is a heuristic approach, it works satisfactorily in practice. 
    
    \noindent {\bf Encoder Network:} 
    The goal of the encoder network $\textnormal{ENC}_{\boldtheta_e}(\cdot):\mathbbR^K\rightarrow\mathbb{R}^h$ is to obtain a low-dimensional hidden representation $\boldh \in \mathbbR^h$ from the output of the selection layer $\boldx^\mathcal{S}$. The encoder network is expressed as follows:
    \begin{equation}
        \textnormal{ENC}_{\boldtheta_e}(\boldx^\mathcal{S}) = \sigma(\boldW_{L_e}^{(e)}\sigma(\cdots \boldW_2^{(e)}\sigma(\boldW_1^{(e)}\boldx^\mathcal{S})\cdots),
    \end{equation}
    where $\boldx^\calS = \textnormal{SEL}(\boldx)$ denotes the output of the selection layer, $\boldtheta_e = \{\boldW_{\ell}^{(e)}\}_{\ell = 1}^{L_e}$ the weight matrix, $L_e$ the number of layers in the encoder network, and $\sigma(\cdot)$ an activation function.
    
    \noindent {\bf Classifier Network:}
    The classifier network $f_{\boldtheta_c}(\cdot):\mathbbR^h\rightarrow\mathcal{Y}$ predicts the final output from the hidden representation $\boldh = \textnormal{ENC}_{\boldtheta_e}(\boldx^\mathcal{S})$ as follows:
    \begin{equation}
      f_{\boldtheta_c}(\boldh) = \textnormal{softmax}(\boldW_{L_y}^{(y)}\sigma(\cdots \boldW_2^{(y)}\sigma(\boldW_1^{(y)}\boldh)\cdots),
    \end{equation}
    where $\boldtheta_c = \{\boldW_{\ell}^{(y)}\}_{\ell = 1}^{L_y}$, and  $L_y$ denotes the number of layers in the classifier network.

    \noindent {\bf Decoder Network:} Generally, a decoder function is employed to reconstruct the original output. However, in this paper, the decoder function $\textnormal{DEC}_{\boldtheta_d}(\cdot):\mathbb{R}^h\rightarrow\mathbb{R}^{h'}$ computes another hidden representation $\widetilde{\boldh} \in \mathbbR^{h'}$  
    and defines the last reconstruction layer separately. The decoder function is defined as follows:
    \begin{equation}
        \textnormal{DEC}_{\boldtheta_d}(\boldh) = \sigma(\boldW_{L_d}^{(d)}\sigma(\cdots \boldW_2^{(d)}\sigma(\boldW_1^{(d)}\boldh)\cdots).
    \end{equation}
    where $\boldh = \textnormal{ENC}_{\boldtheta_e}(\boldx^\mathcal{S})$, $\boldtheta_d = \{\boldW_{\ell}^{(d)}\}_{\ell = 1}^{L_d}$, and $L_d$ denotes the number of layers in the decoder network.

    \noindent {\bf Reconstruction Layer:} To reconstruct the original high-dimensional feature $\boldx$, it must have $O(dh')$ parameters and depend on the dimension $d$. Thus, in a manner similar to the selection layer, we use a tiny network to predict the model parameters. The reconstruction layer is expressed as follows: 
    \begin{equation}
        \textnormal{REC}_{\boldtheta_r}(\widetilde{\boldh}) =  \boldW^{(r)}\widetilde{\boldh},
    \end{equation}
    where $\widetilde{\boldh} = \textnormal{DEC}_{\boldtheta_d}(\boldh)$, $\boldtheta_r = \boldW^{(r)} \in \mathbbR^{d \times h'}$, and $[{\boldW^{(r)}}^\top]_j=\boldvarphi_{\boldomega_r}(\boldphi(\boldu_j))$ denotes the virtual weights of the $j^{th}$ row in the reconstruction layer. The tiny network $\boldvarphi_{\boldomega_r}(\cdot):\mathbbR^b\rightarrow\mathbbR^{h'}$ is trained on $\boldphi(\boldu_j) \in \mathbbR^b$ to predict the weights that connect the $j^{th}$ row of the reconstruction layer to all the $h'$ neurons of the last layer of the decoder network. In this paper, we use $[{\boldW^{(r)}}^\top]_j = \text{tanh}(\boldW_{\boldomega_r} \boldphi(\boldu_j))$, where $\boldW_{\boldomega_r} \in \mathbbR^{h' \times b}$ is the model parameter for the tiny network.

\begin{figure*}[t]
        \centering
        \begin{tabular}{c c}
        \includegraphics[width=0.24\linewidth,keepaspectratio]{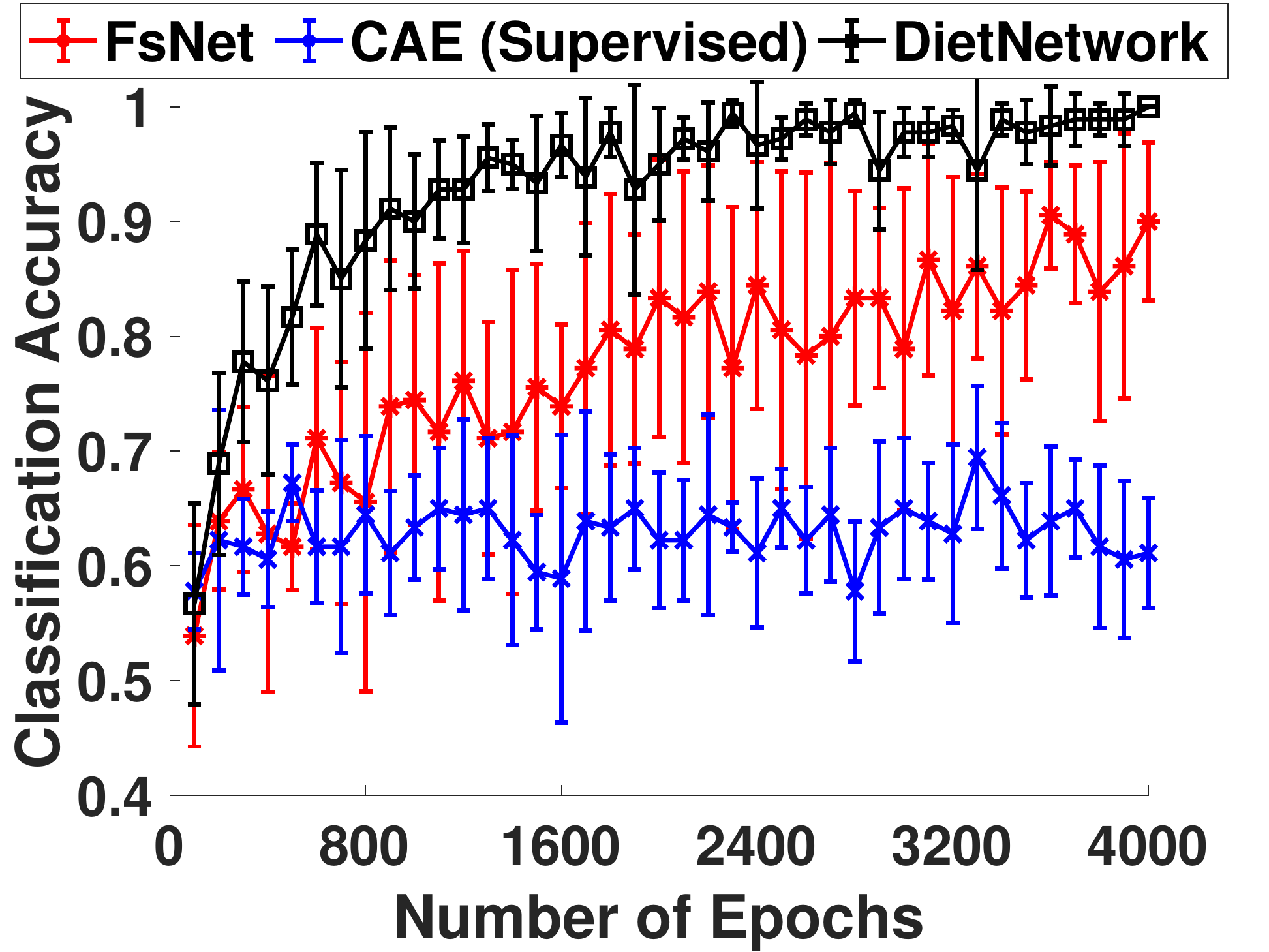}
        \includegraphics[width=0.24\linewidth,keepaspectratio]{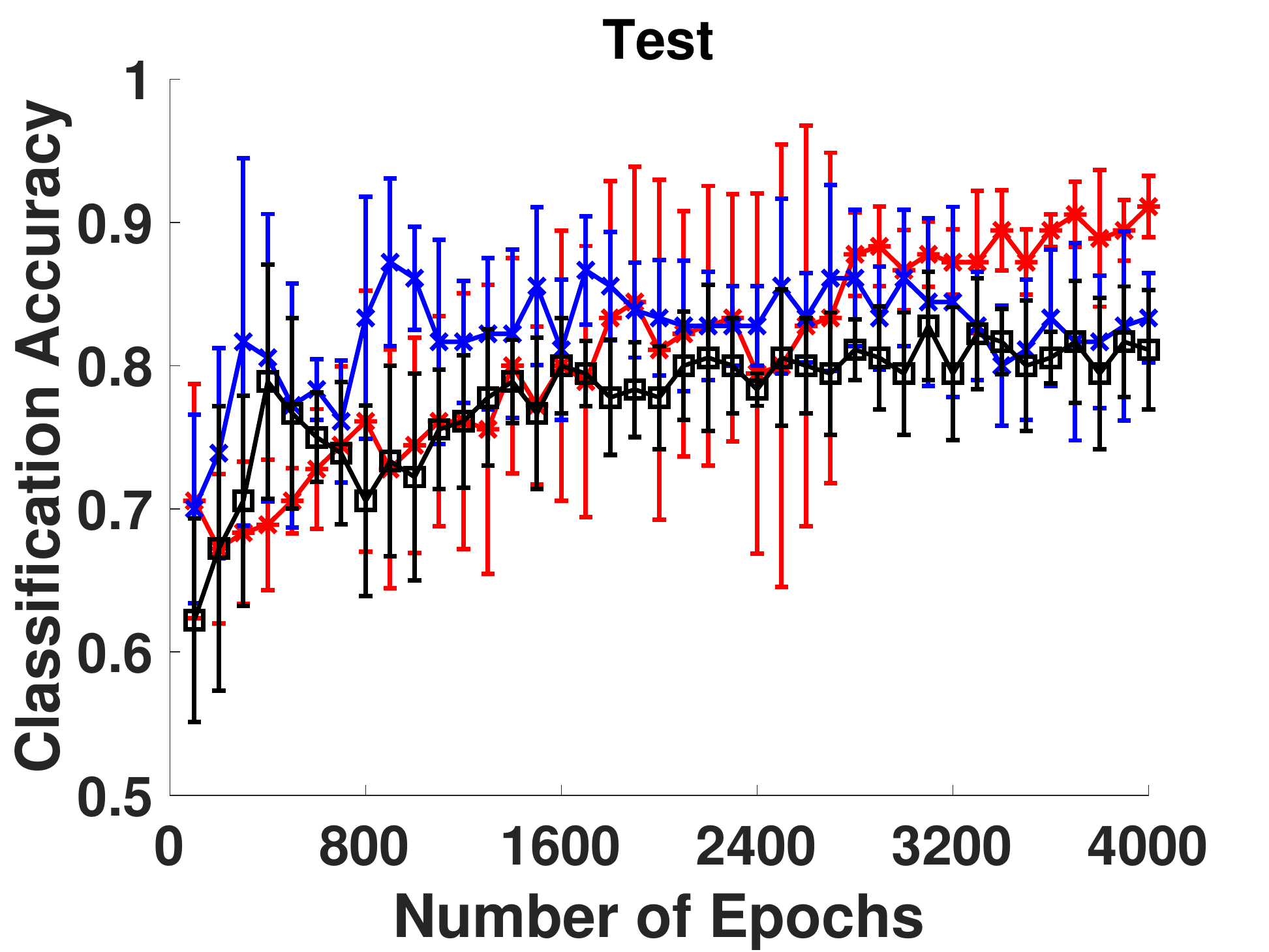}&
        \includegraphics[width=0.24\linewidth,keepaspectratio]{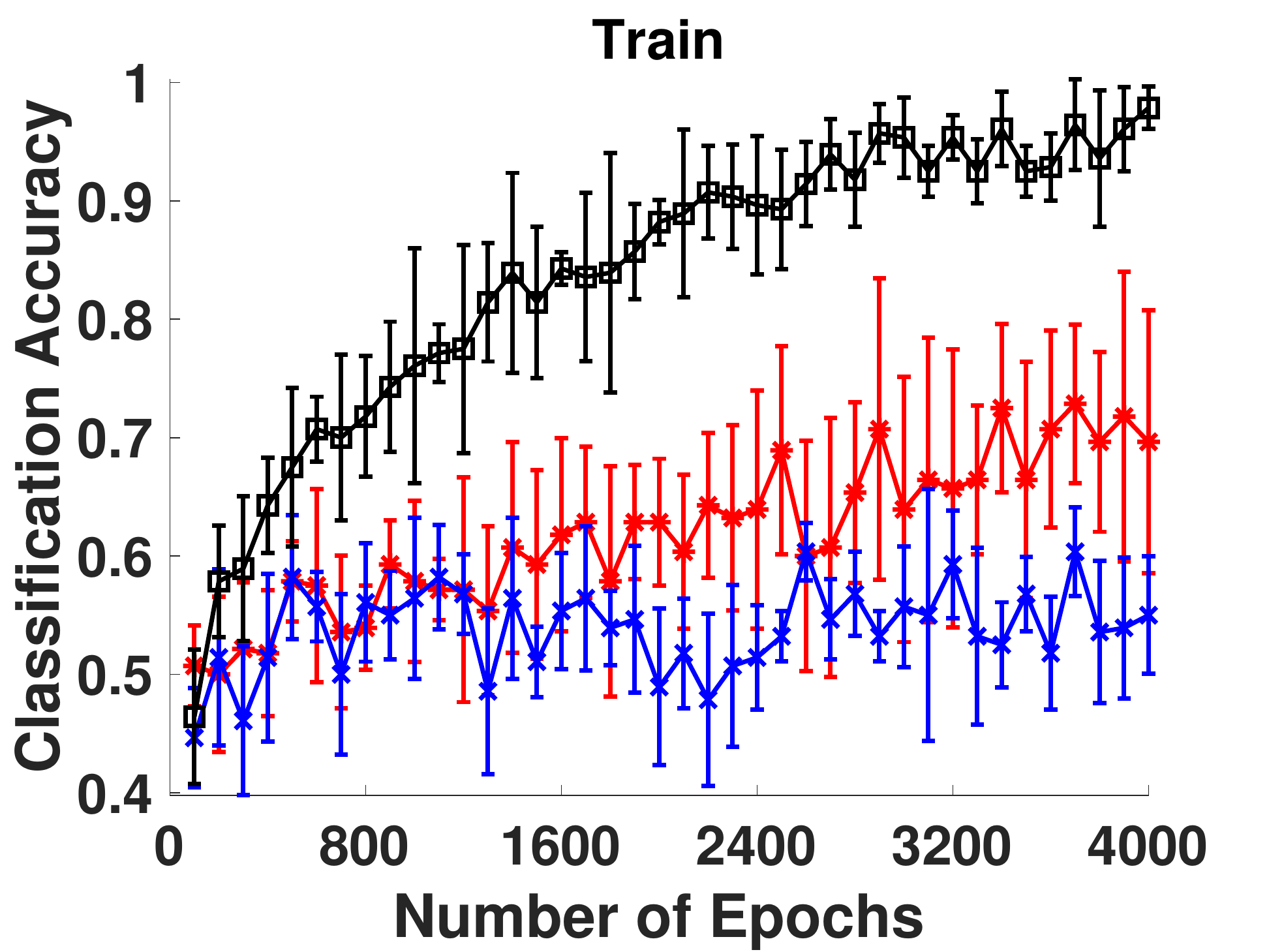}
        \includegraphics[width=0.24\linewidth,keepaspectratio]{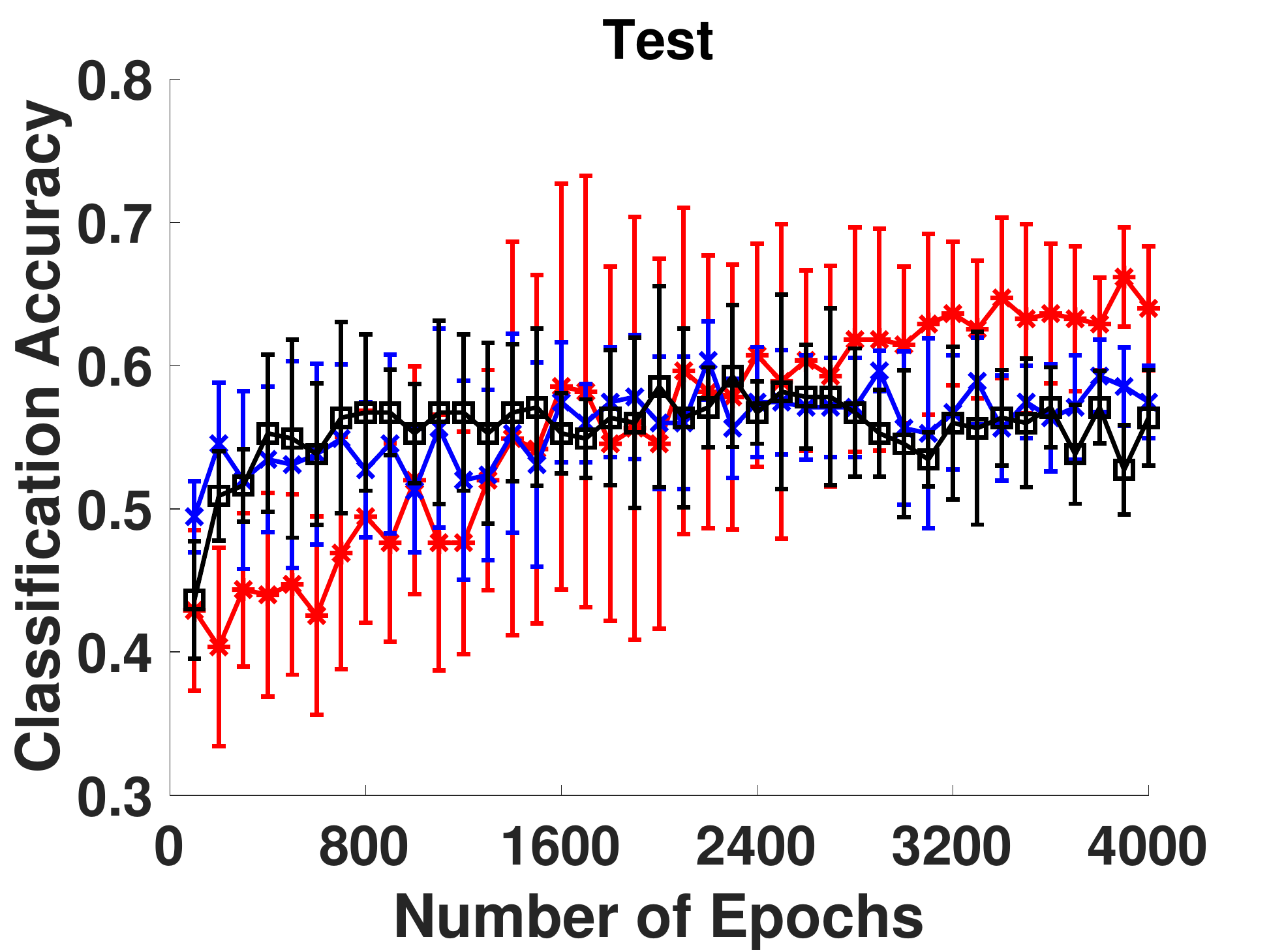}\\
        (A) ALLAML & (B) CLL\_SUB\\
        \end{tabular}
        \vspace{-.15in}
        \caption{Comparison among FsNet, supervised CAE, Diet Network, and SVM for mean training and testing accuracies over the epochs. For the neural-network-based approaches, we set the model parameters to $b = 10$ and $K = 10$. (See all the experimental results in Figure \ref{fig:allacc_full}). }
        \label{fig:allacc}
        \vspace{-.15in}
    \end{figure*}
    
    \begin{figure*}[!t]
        \centering
        \begin{tabular}{c c c}
            \includegraphics[width=0.3\linewidth,keepaspectratio]{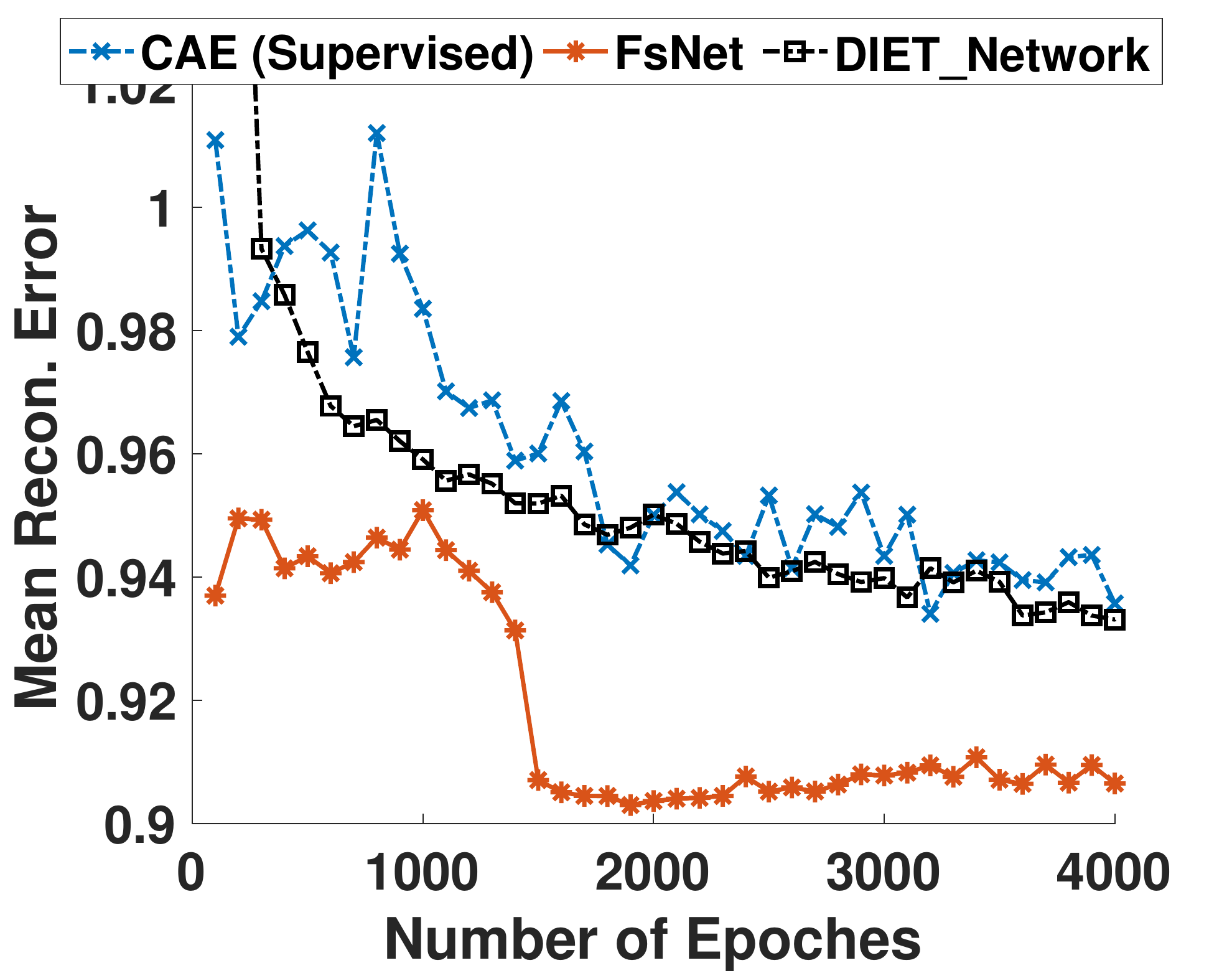}&
            \includegraphics[width=0.3\linewidth,keepaspectratio]{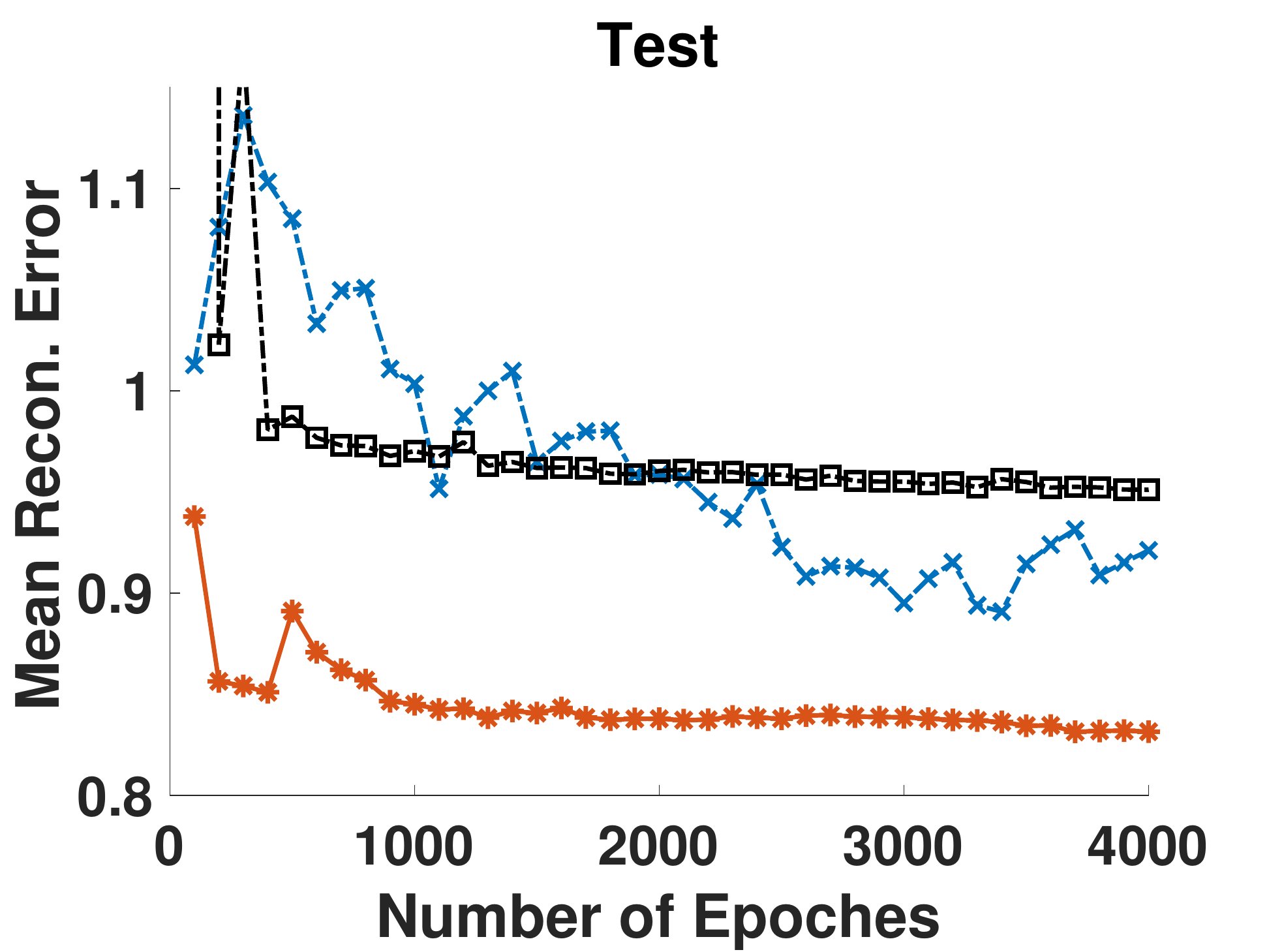}&
            \includegraphics[width=0.3\linewidth,keepaspectratio]{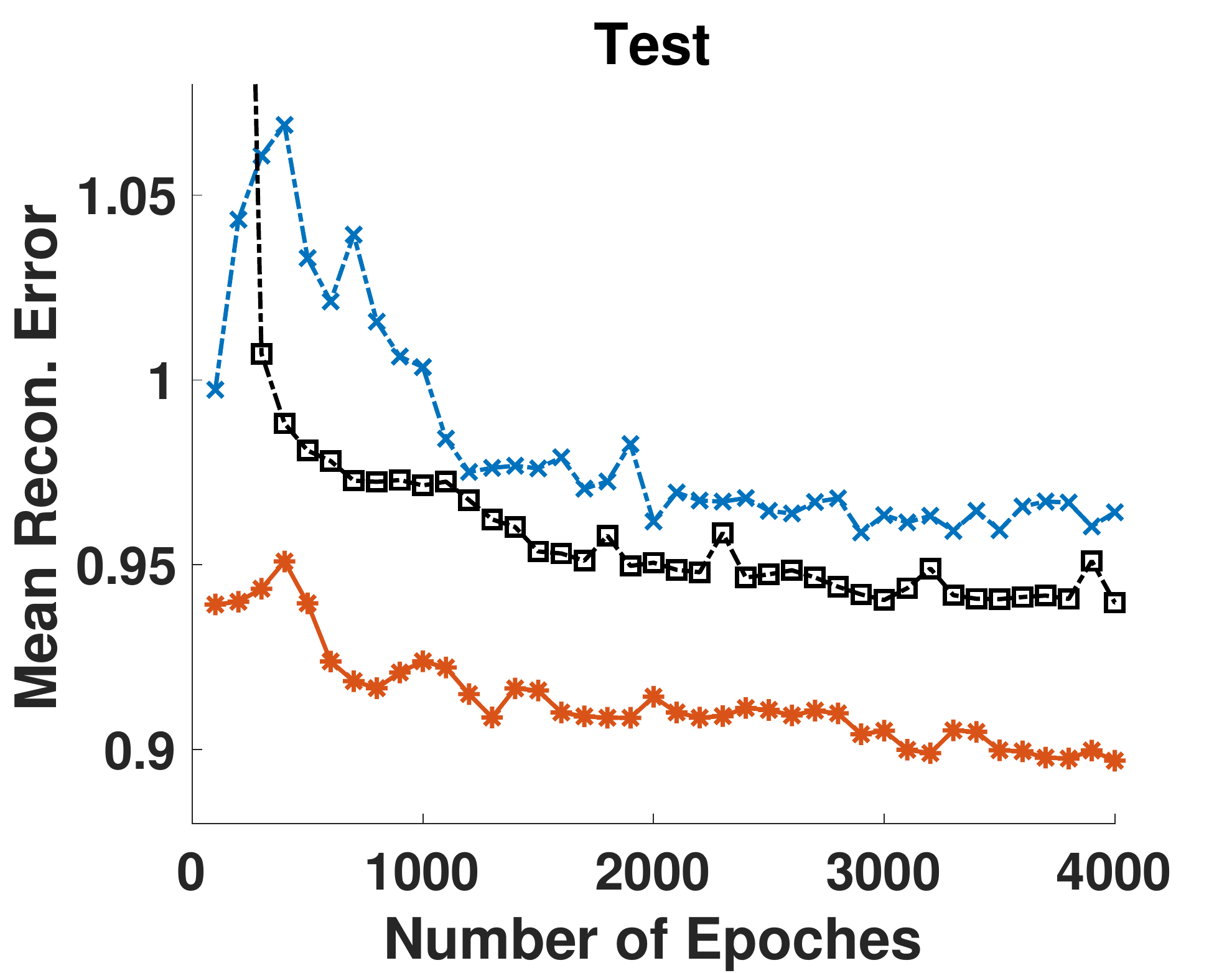}\\
            (A) ALLAML & (B) CLL\_SUB & (C) GLI\_85\\
        \end{tabular}
        \vspace{-.15in}
        \caption{Comparison among the proposed FsNet and existing supervised CAE approaches in terms of the mean test reconstruction error over the epochs. (See the Supplementary material for all the data results). }
        \label{fig:allrecon}
    \end{figure*}

\section{Empirical Evaluation}
    Here, we compare FsNet with several baselines using benchmark and the real metagenome dataset. 
    \subsection{Setup}
    We compared FsNet with CAE~\cite{Balin2019icml}, which is a unsupervised, neural-network-based, feature selection method, Diet Networks \cite{Romero2017iclr}, HSIC Lasso \cite{yamada2014high,Yamada2018tkde,10.1093/bioinformatics/btz333}, and mRMR \cite{Peng2005pami}. Notably, CAE and HSIC Lasso are state-of-the-art, nonlinear feature selection methods, which are deep and shallow, respectively. FsNet and CAE \cite{Balin2019icml} were run on a Linux server with an Intel Xeon CPU Xeon(R) CPU E5-2690 v4 @ 2.60 GHz processor, 256 GB RAM, and NVIDIA P100 graphics card. HSIC Lasso \cite{yamada2014high} and mRMR \cite{Peng2005pami} were executed on a Linux server with an Intel Xeon CPU E7-8890 v4 2.20 GHz processor and 2 TB RAM.
    
    For FsNet and CAE, we conducted experiments on all the datasets using a fixed architecture, defined as $[d\rightarrow K\rightarrow 64\rightarrow 32\rightarrow 16 (\rightarrow |\mathcal{Y}|)\rightarrow 32\rightarrow 64\rightarrow d]$, where $d$ and $|\mathcal{Y}|$ are data dependent, and $K \in \{10, 50\}$. Each hidden layer uses the \textit{leakyReLU} activation function and \textit{dropout} regularization with a dropout rate of $0.2$. We implemented FsNet in \textit{keras} and used the \textit{RMSprop} optimizer for all the experiments. For the regularization parameter $\lambda$, we used $\lambda=1$ for all the experiments. We performed the experiments with $4000$ epochs at a learning rate of $\eta=10^{-3}$, initial temperature $\tau_0=10$, and end temperature $\tau_E=0.01$ in the annealing schedule for all the experiments.

  \subsection{Benchmark Dataset}
   We used six high-dimensional datasets from biological classification problems\footnote{Publicly available at \url{http://featureselection.asu.edu/datasets.php}}. Table~\ref{tab:datasets} in the supplementary material lists the relevant details of these datasets.
   The performance was evaluated on the basis of four parameters: classification accuracy, reconstruction error, mutual information between the selected features, and model size. Because neither HSIC Lasso nor mRMR could directly classify the samples, we used a support vector machine (SVM) (with a radial basis function) trained on the selected features. As CAE is an unsupervised method, we added a softmax layer to its loss function to ensure a fair comparison; the resulting model is henceforth referred to as supervised CAE. Because RMSprop is a stochastic optimizer, all the results reported are the means of $20$ runs on random splits of the datasets.

    \begin{table*}[t]
    \centering 
    \caption{Comparison of the mean testing accuracy among FsNet, supervised CAE, HSIC Lasso (HSIC), and mRMR with $K = 10$ and $K = 50$. Moreover, we report SVM and Diet Networks. $^\ast$ The pyMRMR package, which is a wrapper of the original code, returns a memory error, and we could not execute the models on these datasets.  \label{tab:allacc}}
  \begin{tabular}{lcccc|cccc|cc} 
  \hline
            & \multicolumn{4}{c}{$K = 10$} & \multicolumn{4}{c}{$K = 50$} & \multicolumn{2}{c}{All features}\\ 
    Dataset & FsNet & CAE & HSIC & mRMR  & FsNet & CAE & HSIC & mRMR & SVM & Diet-net \\ \hline
    ALLAML       & \textbf{0.911} & 0.833 & 0.899 &  0.848 & 0.922 & \textbf{0.936} & 0.917 & { 0.919} & 0.819 & 0.811\\ 
    CLL\_SUB     & \textbf{0.640} & 0.575 & 0.604 & N/A$^\ast$ & 0.582 & 0.556 &{\bf 0.680} & N/A$^\ast$& 0.569 & 0.564\\ 
    GLI\_85      & 0.874 & \textbf{0.884} &  0.831 & N/A$^\ast$ & 0.795 & 0.822 & \textbf{0.829} & N/A$^\ast$ & 0.759 & 0.842\\ 
    GLIOMA       & \textbf{0.624} & 0.584 & 0.595 & 0.564 & 0.624 & 0.604 & 0.672 & {\bf 0.693}  & 0.628 & 0.712\\ 
    Prostate\_GE & 0.871 & 0.835 & {\bf 0.924} & 0.871 & 0.878 & 0.884 & 0.926 & {\bf 0.933} & 0.846 & 0.753\\ 
    SMK\_CAN     & \textbf{0.695} & 0.680 & 0.660 & 0.620 & 0.641 & 0.667 & {\bf 0.684} & 0.668 & 0.699 & 0.665\\ \hline
  \end{tabular}
  \vspace{-.2in}
\end{table*}

    \noindent {\bf Classification accuracy:} Figure \ref{fig:allacc} compares the training and testing behaviors of FsNet and supervised CAE for embedding size $b=10$ and number of selected features $K=10$. The results across the datasets show that FsNet can learn better than supervised CAE owing to its reduced number of parameters. The classification performance of FsNet for $10$ selected features is consistently superior to that of the SVM and Diet Networks for all the features across all the datasets. Similarly, the comparable performances of the proposed FsNet for $10$ selected features and Diet-Network with all the features across the datasets illustrate that using a concrete random variable for the continuous relaxation of the discrete feature selection objective does not significantly change the objective function. Additionally, the correlation between the testing and training accuracies of FsNet demonstrates its generalization capability in comparison to supervised CAE, which seems to be overfitted under such high-dimensional data with a limited number of samples. 
    
    Table~\ref{tab:allacc} presents the testing accuracies of the feature selection methods for various numbers of features selected on the six datasets. The experiments show that FsNet performs consistently better than supervised CAE, HSIC Lasso, mRMR, and Diet Networks for $K = 10$. However, the performance of neural-network-based models deteriorates when the number of features $K$ increases. This is because as the number of parameters increases, the training of the model becomes increasingly difficult. Overall, FsNet tends to outperform the baselines even when the number of selected features is small ($K = 10$), and this is a satisfactory property of FsNet.

     The selected features are highly predictive of the target variable. However, they represent the rest of the features in the dataset, as can be seen from the reconstruction error introduced in producing the original features from selected features (see Figure~\ref{fig:allrecon}). FsNet achieves a more competitive reconstruction error than supervised CAE and Diet Network on all the datasets.

\begin{wraptable}[12]{r}[5mm]{80mm}
\vspace{-.2in}
    \centering 
    \caption{Model-size comparison between supervised CAE and FsNet $^{\text{\ref{ftn:modelsize}}}$ (in KBs) at $K = 10$. Because FsNet predicts the model parameter by using a fixed-sized neural network, its model size is the same for all the datasets. \label{tab:modelsize}}
    \small
  \begin{tabular}{lccc} 
  \hline
    Dataset & FsNet & CAE & Compression ratio \\ \hline
    ALLAML       & 108 & 4280 & 39.6  \\ 
    CLL\_SUB     & 108 & 6748 & 62.5 \\ 
    GLI\_85      & 108 & 13160 &121.9 \\ 
    GLIOMA       & 108 & 2704 & 25.0 \\ 
    Prostate\_GE & 108 & 3600 & 33.3\\ 
    SMK\_CAN     & 108 & 11820 & 109.4\\ \hline
  \end{tabular}
\end{wraptable}
    \noindent {\bf Model-size comparison:} 
    The number of parameters in the selection layer of supervised CAE is $O(dK)$, whereas in FsNet, the weight-predictor network of the selection layer has $O(bK)$ parameters. Similarly, the number of parameters in the reconstruction layer of supervised CAE is $O(dh')$, whereas in FsNet, the weight-predictor network of the reconstruction layer has $O(bh')$ parameters. The model compression ratio (CR) for FsNet with respect to supervised CAE is  $CR=\frac{|\theta_s|+|\theta_r|+s}{|\omega_s|+|\omega_r|+s}=\frac{dh+h'd+s}{bh+h'b+s}=O\left(\frac{d}{b}\right),$
    where $s=|\theta_e|+|\theta|+|\theta_d|$ denotes the number of parameters in the rest of the network. Thus, FsNet has $\approx\frac{d}{b}$ times fewer parameters than supervised CAE.
    
    Table~\ref{tab:modelsize} lists the model sizes\footnote{\label{ftn:modelsize}Model size figures are the size of the \textit{keras} model on the disk.} in kilobytes (KBs) for FsNet and supervised CAE. The results show that FsNet can significantly reduce its model size according to the number of selected features ($K$) and size of the feature embedding ($b$). FsNet compresses the model size by 25--122 folds in comparison to supervised CAE. This reduction in the model size of FsNet is due to the use of tiny weight-predictor networks in the fat selection and reconstruction layers.

    \begin{wrapfigure}[11]{r}[5mm]{90mm}
  \centering
  \vspace{-.2in}
  \includegraphics[keepaspectratio,width=70mm]{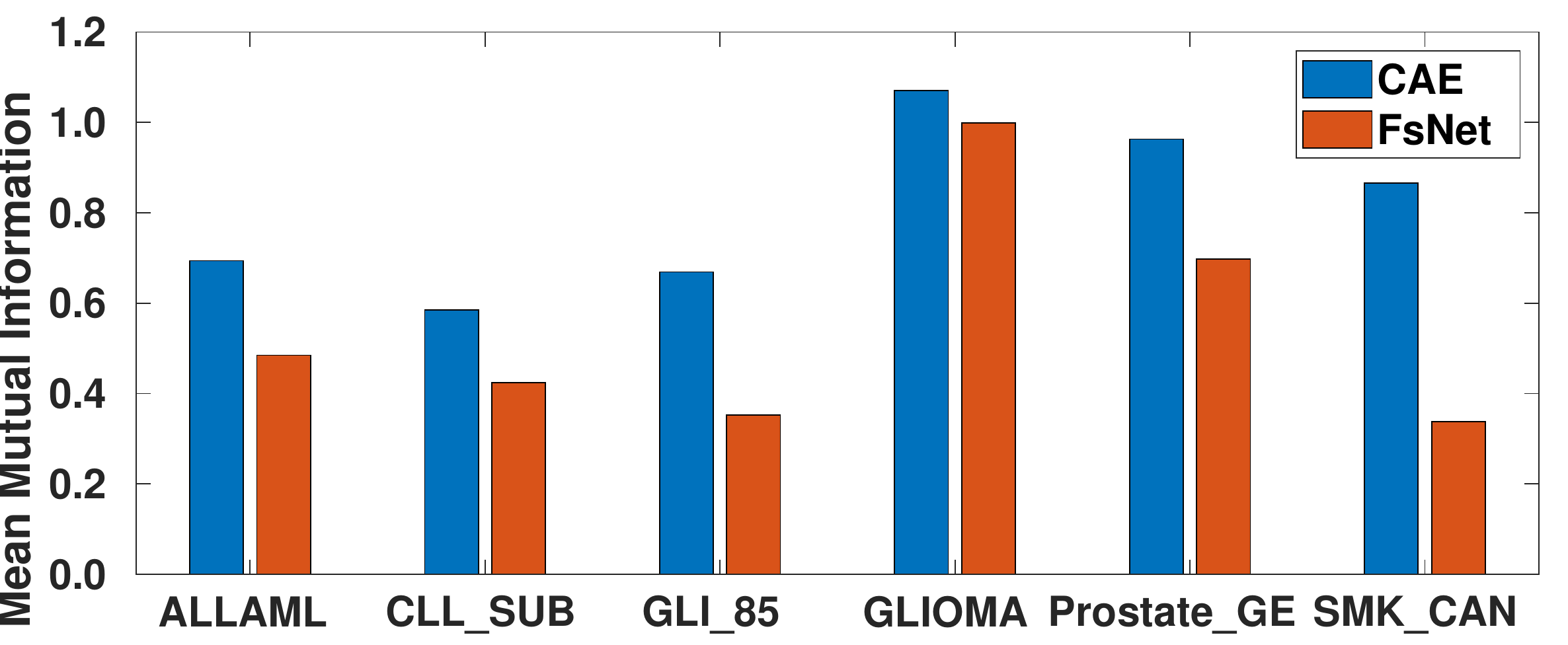}
  \caption{Comparison in terms of the average mutual information between the features selected by CAE and FsNet, respectively. The lower, the better.\label{fig:mi}}
\end{wrapfigure}
    
    \noindent {\bf Minimum redundancy:} The minimum redundancy criterion is important to measure the usefulness of the selected features. According to this criterion, the selected features should have minimum dependencies between themselves. We used the average mutual information between all the pairs of the selected features to compare the validity of the features selected by FsNet and CAE, respectively. The average mutual information is defined as follows: $\hat{I}(\mathcal{S})=\frac{2}{K(K-1)}\sum_{i,j\in\mathcal{S},j>i}I(X_i,X_j)$, where $I(X_i,X_j)$ denotes the mutual information between features $i$ and $j$ in the selected set $\calS$.
    
    As shown in Figure~\ref{fig:mi}, compared with CAE, the average mutual information between the features selected by FsNet is significantly lower on all the datasets. This shows that compared with CAE, FsNet more effectively selects the features with minimum redundancy owing to the use of \textit{unique\_argmax} functions in the selection layer. 

    \begin{wraptable}[9]{r}[0mm]{80mm}
\vspace{-.2in}
    \centering 
    \caption{Classification accuracy of different methods on the metagenome dataset on inflammatory bowel disease.}
    \label{tab:metagenome_accruacy}
    \small
  \begin{tabular}{lcc} 
  \hline
                        & \multicolumn{2}{c}{Accuracy} \\
    Method           & $K = 50$ & $K = 100$ \\
    \hline
    FsNet 	       & 0.907 $\pm$ 0.017 & 0.917 $\pm$ 0.014 \\ 
    CAE           & 0.750 $\pm$ 0.065 & 0.700 $\pm$ 0.067 \\
    HSIC Lasso (B=10)  & 0.931 $\pm$ 0.006 & 0.931 $\pm$ 0.006 \\
    HSIC Lasso (B=20)  & 0.942 $\pm$ 0.002 & 0.943 $\pm$ 0.002 \\
    mRMR            	& 0.944 $\pm$ 0.002 & 0.933 $\pm$ 0.002 \\
    \hline
    SVM 	            & \multicolumn{2}{c}{0.876 $\pm$ 0.002} \\
    Diet-networks 	    & \multicolumn{2}{c}{0.902 $\pm$ 0.023} \\
    \hline
  \end{tabular}
\end{wraptable}    
    

\subsection{Application to inflammatory bowel disease}
We studied a metagenome dataset \cite{lloyd2019multi}, which contains information regarding the gut bacteria of 359 healthy individuals and 958 patients with inflammatory bowel disease. Specifically, 7\,547 features are KEGG orthology accession numbers, which represent molecular functions to which reads from the guts of samples guts are mapped. We included three additional features: age, sex, and race.

We selected 50 or 100 features on this dataset using FsNet, CAE, HSIC Lasso, and mRMR. For HSIC Lasso, as the number of samples was high, we employed  the block HSIC Lasso \cite{10.1093/bioinformatics/btz333}, where $B$ denotes the tuning parameter of the block HSIC Lasso, and $B = n$ is equivalent to the standard HSIC Lasso \cite{yamada2014high}.  The Diet-networks outperformed SVM. The proposed FsNet could achieve better performance than Diet-networks with significantly less number of features, while CAE failed to train the model due to the large number of model parameters. Moreover, the model compression ratio between FsNet and CAE is 21.41, and thus we conclude that FsNet can obtain preferable performance with much less number of parameters for high-dimensional data.

\section{Conclusions}
    We proposed FsNet, which is an end-to-end trainable, deep learning-based, feature selection method for high-dimensional data with a small number of samples. FsNet can select unique features by using a concrete random variable. Using weight-predictor functions and a reconstruction loss, it not only required few parameters but also stabilized the model and made it appropriate for training with a limited number of samples. The experiments on several high-dimensional biological datasets demonstrated the robustness and superiority of FsNet for feature selection in the chosen settings. Moreover, we evaluated the proposed FsNet on a real-life metagenome dataset, and FsNet compares favorably with existing shallow models. 

\section*{Acknowledgement}
We acknowledge the iHMP principal investigator(s) whose data we used and the HMP program. These include support from the NIH Common Fund for grant number 8U54HD080784 to Gregory Buck, Jerome Strauss, Kimberly Jefferson (Virginia Commonwealth University), grant number 8U54DK102557 to Ramnik Xavier, Curtis Huttenhower (Broad Institute) and grant number 8U54DK102556 to Michael Snyder (Stanford University) with additional co-funding from the National Center for Complementary and Alternative Medicine (NCCAM), the Office of Research on Women’s Health (ORWH) and the Office of Dietary Supplements (ODS).

    \bibliography{main}

\begin{thebibliography}{10}

\bibitem{Balasubramanian2013aistats}
Krishnakumar Balasubramanian, Bharath~K. Sriperumbudur, and Guy Lebanon.
\newblock Ultrahigh dimensional feature screening via {RKHS} embeddings.
\newblock In {\em {AISTATS}}, 2013.

\bibitem{Balin2019icml}
Muhammed~Fatih Balin, Abubakar Abid, and James~Y. Zou.
\newblock Concrete autoencoders: Differentiable feature selection and
  reconstruction.
\newblock In {\em {ICML}}, 2019.

\bibitem{Chen2015icml}
Wenlin Chen, James~T. Wilson, Stephen Tyree, Kilian~Q. Weinberger, and Yixin
  Chen.
\newblock Compressing neural networks with the hashing trick.
\newblock In {\em {ICML}}, 2015.

\bibitem{10.1093/bioinformatics/btz333}
Héctor Climente-González, Chloé-Agathe Azencott, Samuel Kaski, and Makoto
  Yamada.
\newblock {Block HSIC Lasso: model-free biomarker detection for ultra-high
  dimensional data}.
\newblock {\em Bioinformatics}, 35(14):i427--i435, 07 2019.

\bibitem{Dahl2013icassp}
G.~E. Dahl, J.~W. Stokes, Li~Deng, and Dong Yu.
\newblock Large-scale malware classification using random projections and
  neural networks.
\newblock In {\em {ICASSP}}, 2013.

\bibitem{fan2008sure}
Jianqing Fan and Jinchi Lv.
\newblock Sure independence screening for ultrahigh dimensional feature space.
\newblock {\em Journal of the Royal Statistical Society: Series B (Statistical
  Methodology)}, 70(5):849--911, 2008.

\bibitem{guyon2003introduction}
Isabelle Guyon and Andr{\'e} Elisseeff.
\newblock An introduction to variable and feature selection.
\newblock {\em Journal of machine learning research}, 3(Mar):1157--1182, 2003.

\bibitem{Li2014gpb}
Yixue Li and Luonan Chen.
\newblock Big biological data: Challenges and opportunities.
\newblock {\em Genomics, Proteomics {\&} Bioinformatics}, 12(5):187--189, 2014.

\bibitem{Liao2019ijcai}
Shuangli Liao, Quanxue Gao, Feiping Nie, Yang Liu, and Xiangdong Zhang.
\newblock Worst-case discriminative feature selection.
\newblock In {\em {IJCAI}}, 2019.

\bibitem{Liu2017ijcai}
Bo~Liu, Ying Wei, Yu~Zhang, and Qiang Yang.
\newblock Deep neural networks for high dimension, low sample size data.
\newblock In {\em {IJCAI}}, 2017.

\bibitem{lloyd2019multi}
Jason Lloyd-Price, Cesar Arze, Ashwin~N Ananthakrishnan, Melanie Schirmer,
  Julian Avila-Pacheco, Tiffany~W Poon, Elizabeth Andrews, Nadim~J Ajami,
  Kevin~S Bonham, Colin~J Brislawn, et~al.
\newblock Multi-omics of the gut microbial ecosystem in inflammatory bowel
  diseases.
\newblock {\em Nature}, 569(7758):655--662, 2019.

\bibitem{Lu2018nips}
Yang~Young Lu, Yingying Fan, Jinchi Lv, and William~Stafford Noble.
\newblock {DeepPINK: reproducible feature selection in deep neural networks}.
\newblock In {\em {NeurIPS}}, 2018.

\bibitem{Maddison2017iclr}
C.~J. Maddison, Andriy Mnih, and Yee~Whye Teh.
\newblock The concrete distribution: {A} continuous relaxation of discrete
  random variables.
\newblock In {\em {ICLR}}, 2017.

\bibitem{Vivien2013nature}
Vivien Marx.
\newblock {The big challenges of big data}.
\newblock {\em Nature}, 498(7453), 2013.

\bibitem{Masaeli2010icml}
Mahdokht Masaeli, Glenn Fung, and Jennifer~G. Dy.
\newblock From transformation-based dimensionality reduction to feature
  selection.
\newblock In {\em {ICML}}, 2010.

\bibitem{Ming2019ijcai}
Di~Ming and Chris Ding.
\newblock Robust flexible feature selection via exclusive {L21} regularization.
\newblock In {\em {IJCAI}}, 2019.

\bibitem{Peng2005pami}
Hanchuan Peng, Fuhui Long, and Chris H.~Q. Ding.
\newblock Feature selection based on mutual information: Criteria of
  max-dependency, max-relevance, and min-redundancy.
\newblock {\em {IEEE TPAMI}}, 27(8):1226--1238, 2005.

\bibitem{Romero2017iclr}
Adriana Romero, Pierre~Luc Carrier, et~al.
\newblock Diet networks: Thin parameters for fat genomics.
\newblock In {\em {ICLR}}, 2017.

\bibitem{song2007supervised}
Le~Song, Alex Smola, Arthur Gretton, Karsten~M Borgwardt, and Justin Bedo.
\newblock Supervised feature selection via dependence estimation.
\newblock In {\em Proceedings of the 24th international conference on Machine
  learning}, pages 823--830, 2007.

\bibitem{Tibshirani1996jrss}
Robert Tibshirani.
\newblock Regression shrinkage and selection via the lasso.
\newblock {\em J. Royal Stat. Society}, 58(1):267--288, 1996.

\bibitem{Verikas2002prl}
Antanas Verikas and Marija Bacauskiene.
\newblock Feature selection with neural networks.
\newblock {\em Pattern Recognition Letters}, 23(11):1323--1335, 2002.

\bibitem{Vincent2010jmlr}
Pascal Vincent, Hugo Larochelle, et~al.
\newblock Stacked denoising autoencoders: Learning useful representations in a
  deep network with a local denoising criterion.
\newblock {\em {JMLR}}, 11:3371--3408, 2010.

\bibitem{Wang2014nips}
Qian Wang, Jiaxing Zhang, Sen Song, and Zheng Zhang.
\newblock Attentional neural network: Feature selection using cognitive
  feedback.
\newblock In {\em {NIPS}}, 2014.

\bibitem{Wojcik2019paa}
Piotr~Iwo W{\'{o}}jcik and Marcin Kurdziel.
\newblock Training neural networks on high-dimensional data using random
  projection.
\newblock {\em {PAA}}, 22(3):1221--31, 2019.

\bibitem{yamada2014high}
Makoto Yamada, Wittawat Jitkrittum, Leonid Sigal, Eric~P Xing, and Masashi
  Sugiyama.
\newblock High-dimensional feature selection by feature-wise kernelized lasso.
\newblock {\em Neural computation}, 26(1):185--207, 2014.

\bibitem{Yamada2018tkde}
Makoto Yamada, Jiliang Tang, et~al.
\newblock Ultra high-dimensional nonlinear feature selection for big biological
  data.
\newblock {\em {IEEE TKDE}}, 30(7):1352--1365, 2018.

\bibitem{Yamada2018aistats}
Makoto Yamada, Yuta Umezu, Kenji Fukumizu, and Ichiro Takeuchi.
\newblock Post selection inference with kernels.
\newblock In {\em {AISTATS}}, 2018.

\bibitem{yamada2020feature}
Yutaro Yamada, Ofir Lindenbaum, Sahand Negahban, and Yuval Kluger.
\newblock Feature selection using stochastic gates.
\newblock {\em ICML}, 2020.

\bibitem{Ye2019ijcai}
Xiucai Ye, Hongmin Li, Akira Imakura, and Tetsuya Sakurai.
\newblock Distributed collaborative feature selection based on intermediate
  representation.
\newblock In {\em {IJCAI}}, 2019.

\end{thebibliography}
\bibliographystyle{plain}

\newpage

\section*{Supplementary Materials}
\begin{algorithm}[!htb]
        \caption{Training of FsNet}
        \label{alg:FsNet}
        \textbf{Input}: data matrix $\boldX\in\mathbb{R}^{n\times d}$, output labels $y\in \{1,\cdots,L\})$, $K$ target number of features, encoder network $\textnormal{ENC}_{\boldtheta_e}(\cdot)$, decoder network $\textnormal{DEC}_{\boldtheta_d}(\cdot)$, reconstruction function $\textnormal{REC}_{\boldtheta_r}(\cdot)$, classification network $f_{\boldtheta_c}(\cdot)$, weight prediction networks $\boldvarphi_{\boldomega_s}(\cdot)$ \& $\boldvarphi_{\boldomega_s}(\cdot)$, learning rate $\eta$, 
        start temperature $\tau_0$, end temperature $\tau_E$, and number of epochs $E$\\
        \textbf{Output}: set of selected features $\mathcal{S}$, model parameters $\boldTheta$
        
        \begin{algorithmic}[1] 
            \STATE Initialize $\boldTheta=\{\boldomega_s,\boldtheta_e,\boldtheta_d,\boldomega_r,\boldtheta_c\}$.
            \FOR{$e\in\{1,\cdots,E\}$}
                \STATE Update the temperature $\tau=\tau_0(\tau_E/\tau_0)^{e/E}$
                \STATE $(\bolddelta_{s,1},\cdots\bolddelta_{s,d})\leftarrow(\boldvarphi_{\boldomega_{s}}(\boldphi(\boldu_1))\cdots\boldvarphi_{\boldomega_{s}}(\boldphi(\boldu_d)))$
                \STATE $\boldsymbol{\mu}^{(k)}\leftarrow \textnormal{Concrete}(\boldsymbol{\theta_s}^{(k)},\tau)$ using~(\ref{eq:crv})
                \STATE $\boldM \leftarrow (\boldsymbol{\mu}^{(1)},\cdots,\boldsymbol{\mu}^{(K)})^\top$
                \STATE $\mathcal{S}\leftarrow \textnormal{uargmax}(\boldM^\top)$
                \STATE $\boldh\leftarrow 
                        \begin{cases}
                        \textnormal{ENC}_{\boldtheta_e}(\boldM\boldx_i)&\textnormal{if training},\\
                        \textnormal{ENC}_{\boldtheta_e}(\boldx^\mathcal{S})&{\textnormal{inference}}
                        \end{cases}$
                \STATE $\widehat{{y}}\leftarrow f_{\theta}(\boldh)$
                \STATE $\widetilde{\boldh}\leftarrow \textnormal{DEC}_{\boldtheta_d}(\boldh)$
                \STATE $(\boldtheta_{r}^{(1)},\cdots\boldtheta_{r}^{(d)})\leftarrow(\boldvarphi_{\boldomega_{r}}(\boldphi(\boldu_1))\cdots\boldvarphi_{\boldomega_{r}}(\boldphi(\boldu_d)))$
                \STATE $\widehat{\boldx}\leftarrow \textnormal{REC}_{\boldtheta_r}(\widetilde{\boldh})$
                \STATE Define the loss $L$. 
                \STATE Compute $\nabla_{\boldsymbol\omega_r}L$, $\nabla_{\theta}L$, $\nabla_{\theta_d}L$, and $\nabla_{\theta_e}L$ using backpropagation.
                \STATE Compute $\nabla_{\boldsymbol\omega_s^{(k)}}L$ using reparameterization trick
                \STATE Update $\boldsymbol{\omega}_r\leftarrow\boldsymbol{\omega}_r-\eta\nabla_{\boldsymbol\omega_r}L$,\quad$\boldtheta\leftarrow\boldtheta-\eta\nabla_{\boldtheta}L$,\\\hspace{31pt}$\boldtheta_d\leftarrow\boldtheta_d-\eta\nabla_{\boldtheta_d}L$,\quad$\boldtheta_e\leftarrow\boldtheta_e-\eta\nabla_{\boldtheta_e}L$, and\\\hspace{31pt}$\boldsymbol{\omega}_r^{(k)}\leftarrow\boldsymbol{\omega}_r^{(k)}-\eta\nabla_{\boldsymbol\omega_r^{(k)}}L$
            \ENDFOR
            \STATE \textbf{return} $\mathcal{S},\boldTheta$
        \end{algorithmic}
    \end{algorithm}

\begin{table}[ht]
        \small
        \centering
        \caption{Details of Datasets used in this paper}
        \begin{tabular}{lccc}
            \hline
            \textbf{Dataset} & \textbf{Classes} & \textbf{Sample Size } ($n$) & \textbf{Dimensions} ($d$)  \\
            \hline
            ALLAML & 2 &  72 &	7,129  \\
            CLL\_SUB & 3 &  111 &	11,340  \\
            GLI\_85& 2 &  85 &	22,283  \\
            GLIOMA& 4 &  50 &	4,434  \\
            Prostate\_GE & 2 &  102 &	5,966  \\
            SMK\_CAN & 2 &  187 &	19,993  \\
            \hline
        \end{tabular}
        \label{tab:datasets}
    \end{table}
    
    \begin{figure*}[t]
        \centering
        \begin{tabular}{c c}
        \includegraphics[width=0.24\linewidth,keepaspectratio]{figs/ALLAML_train_acc.pdf}
        \includegraphics[width=0.24\linewidth,keepaspectratio]{figs/ALLAML_test_acc.pdf}&
        \includegraphics[width=0.24\linewidth,keepaspectratio]{figs/CLL_SUB_111_train_acc.pdf}
        \includegraphics[width=0.24\linewidth,keepaspectratio]{figs/CLL_SUB_111_test_acc.pdf}\\
        (A) ALLAML & (B) CLL\_SUB\\
        \includegraphics[width=0.24\linewidth,keepaspectratio]{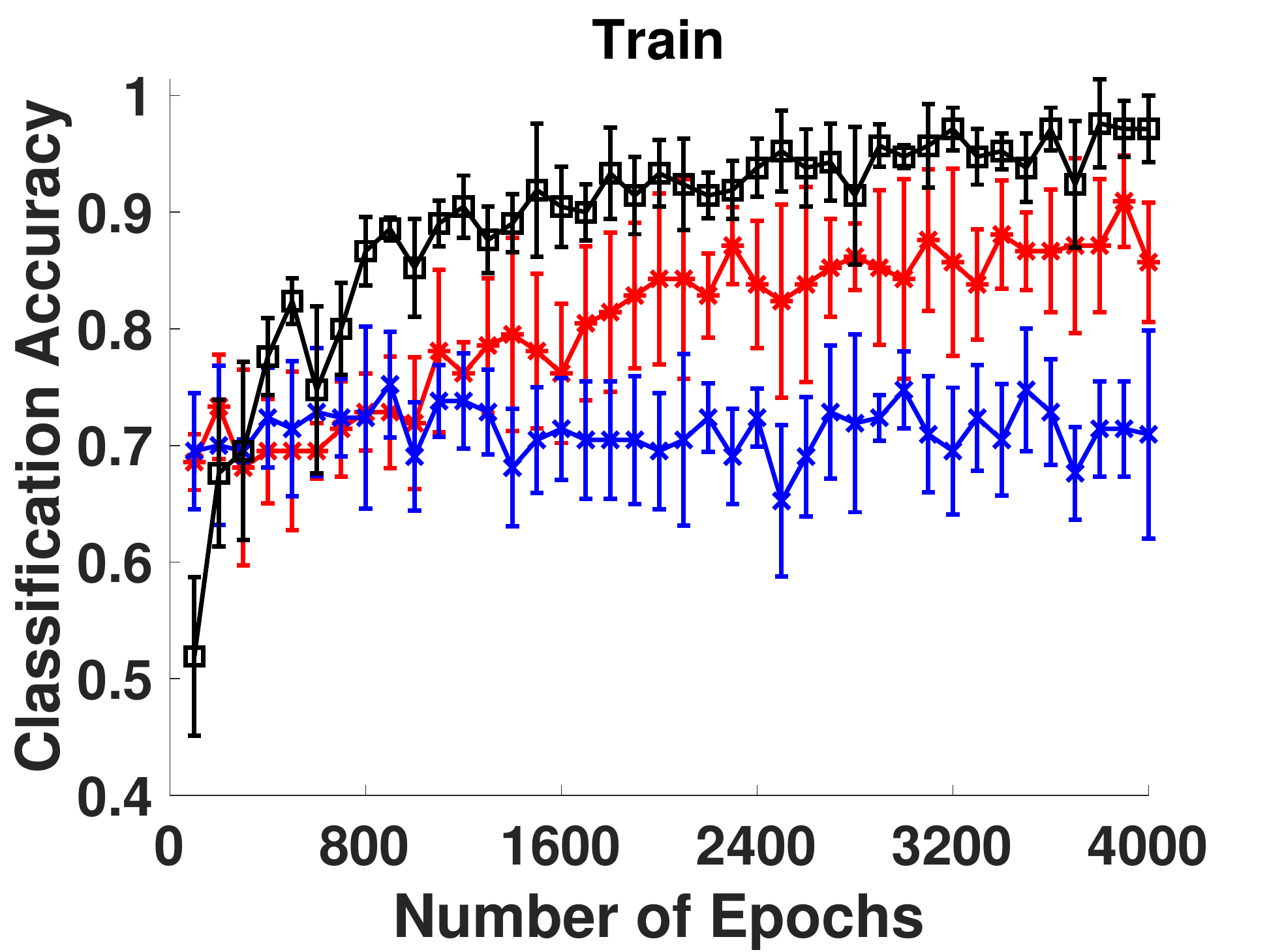}
        \includegraphics[width=0.24\linewidth,keepaspectratio]{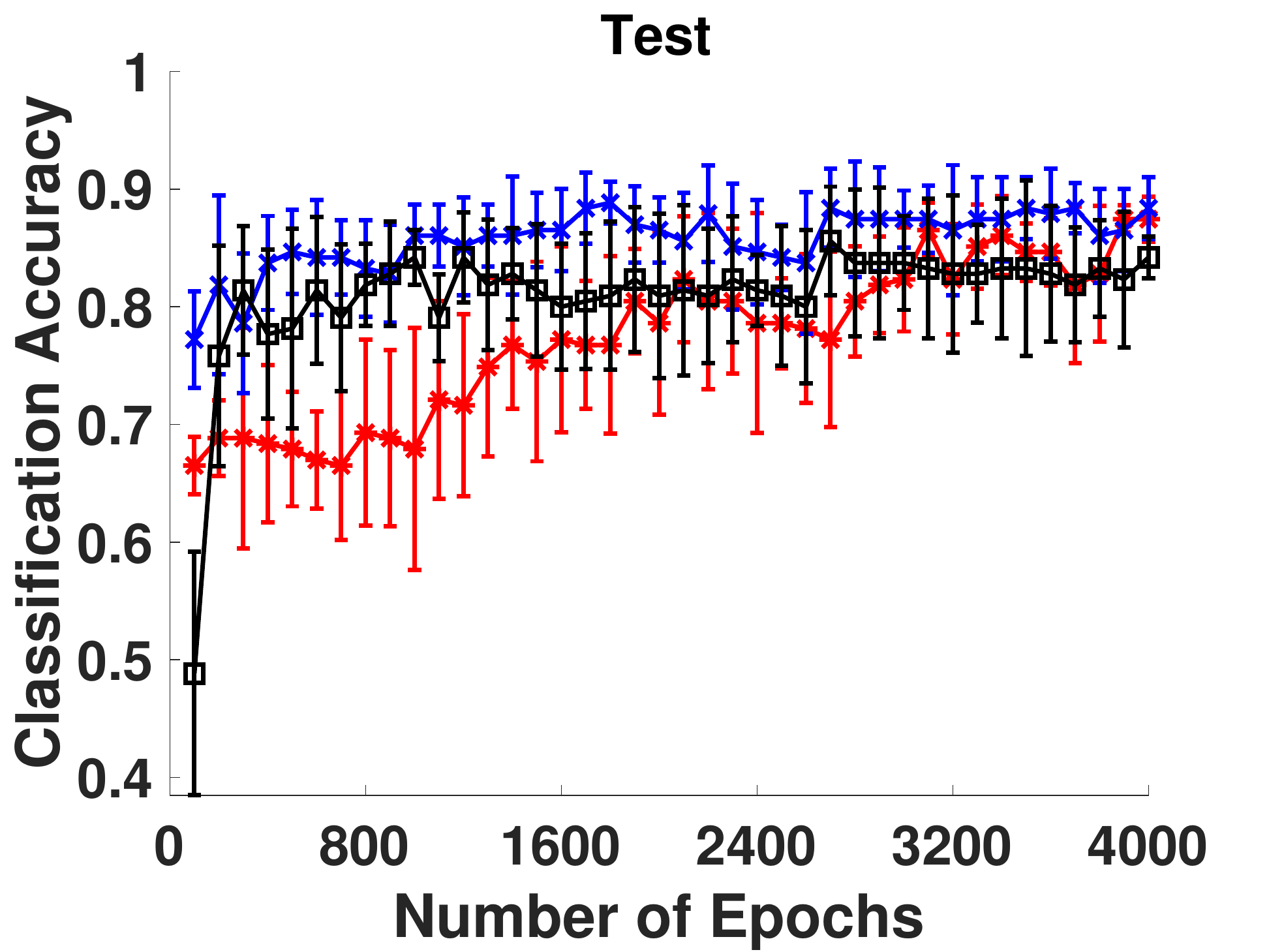}&
        \includegraphics[width=0.24\linewidth,keepaspectratio]{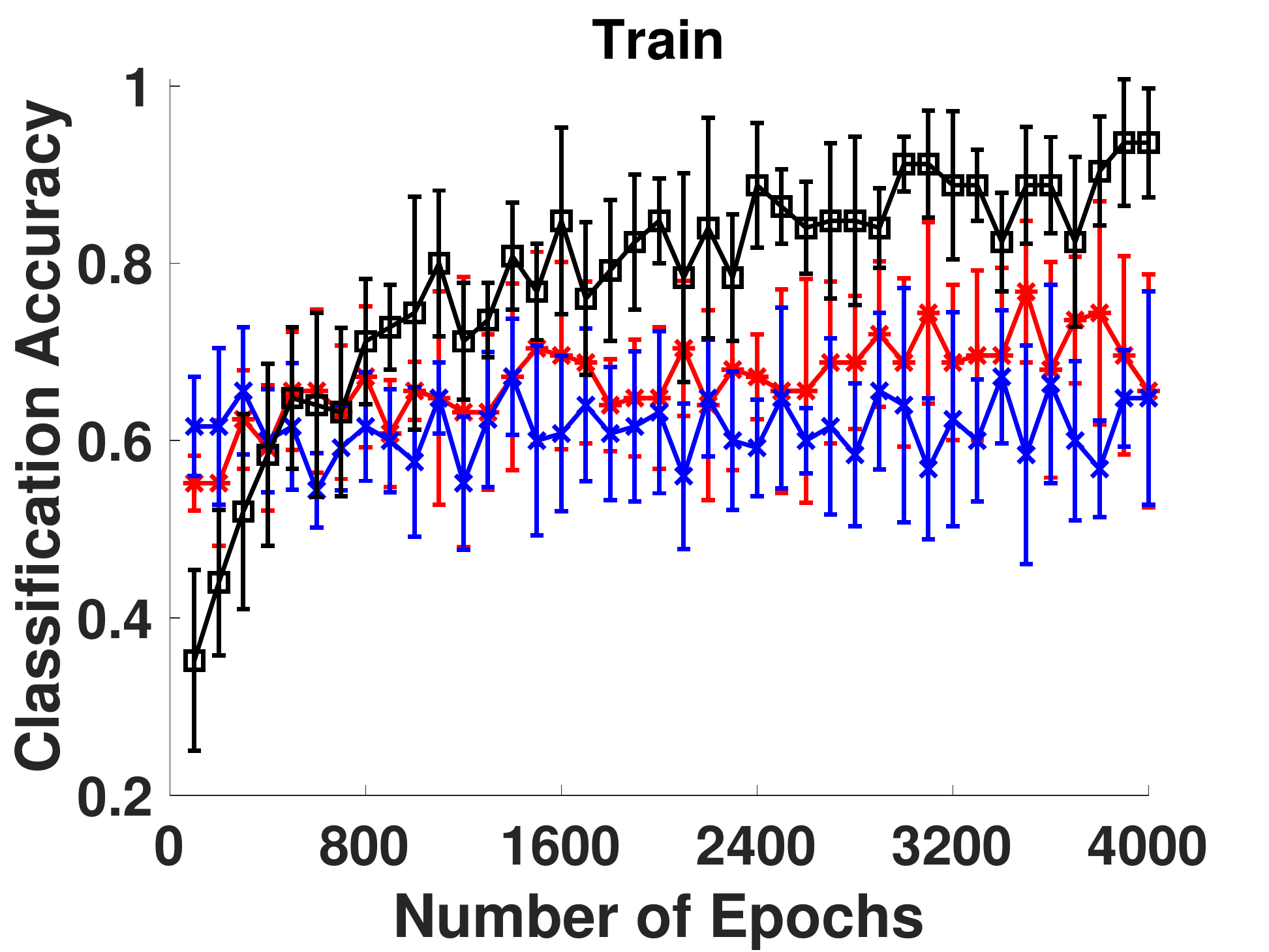}
        \includegraphics[width=0.24\linewidth,keepaspectratio]{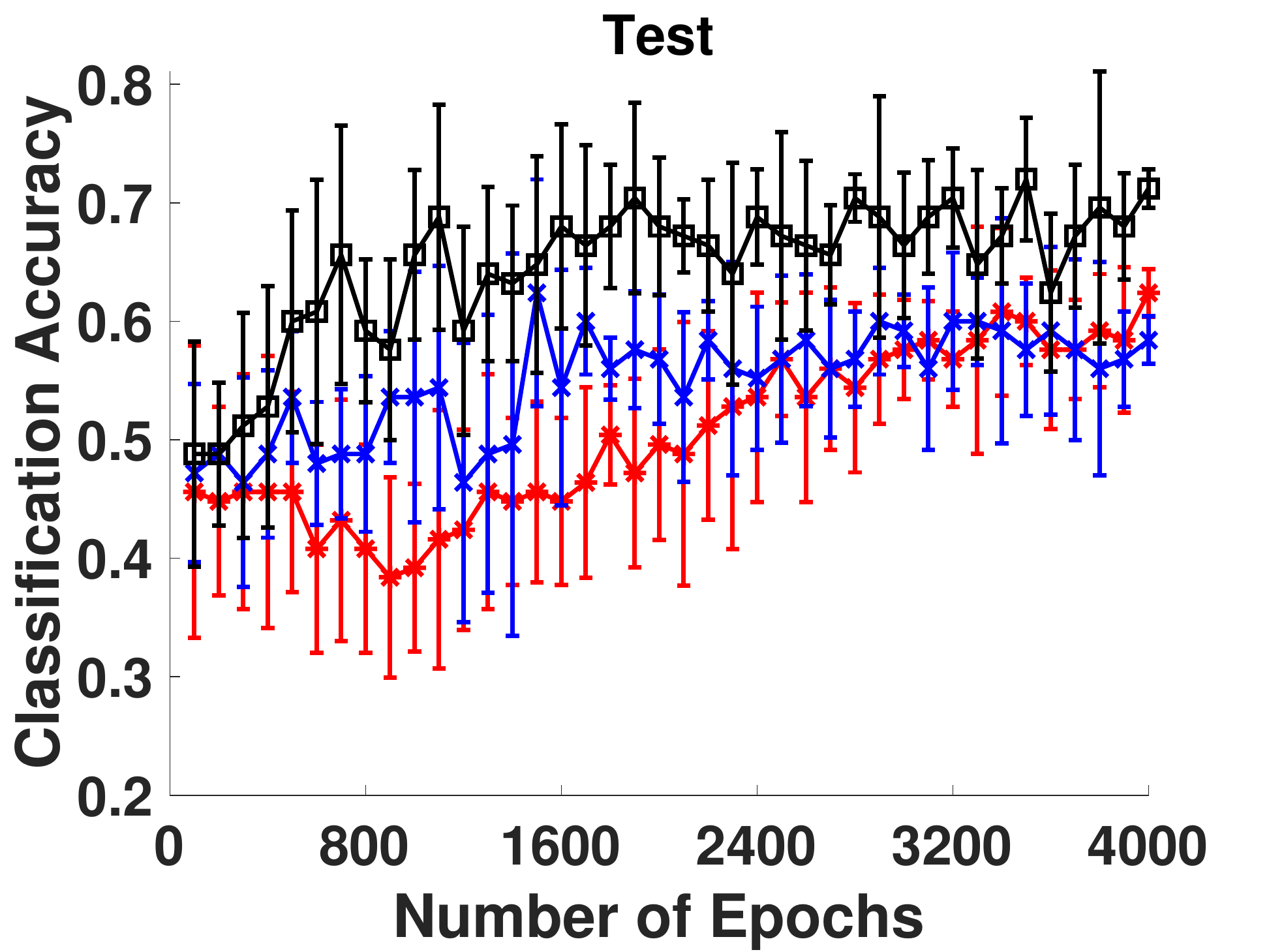}\\
        (C) GLI\_85 & (D) GLIOMA\\
        \includegraphics[width=0.24\linewidth,keepaspectratio]{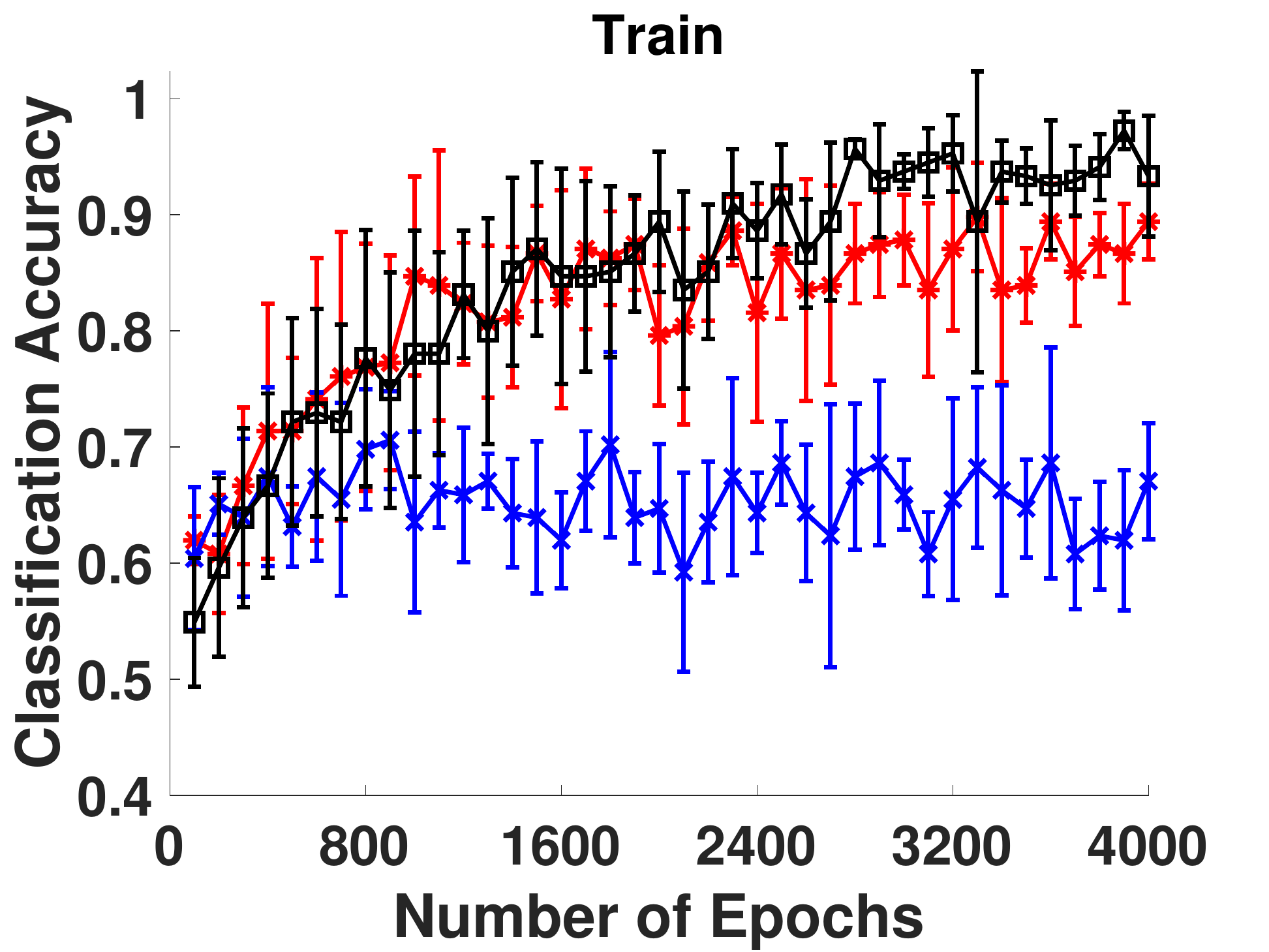}
        \includegraphics[width=0.24\linewidth,keepaspectratio]{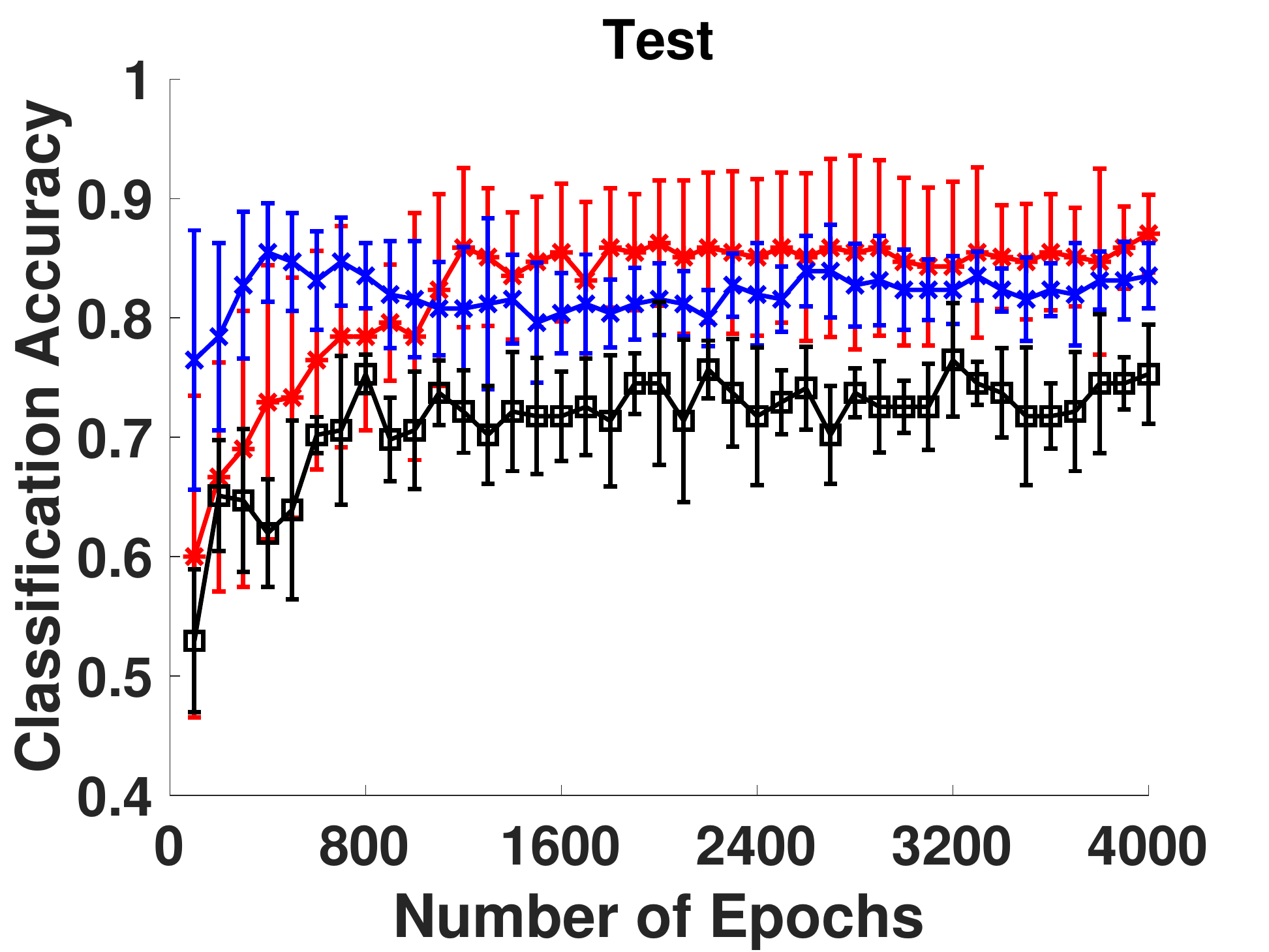}&
        \includegraphics[width=0.24\linewidth,keepaspectratio]{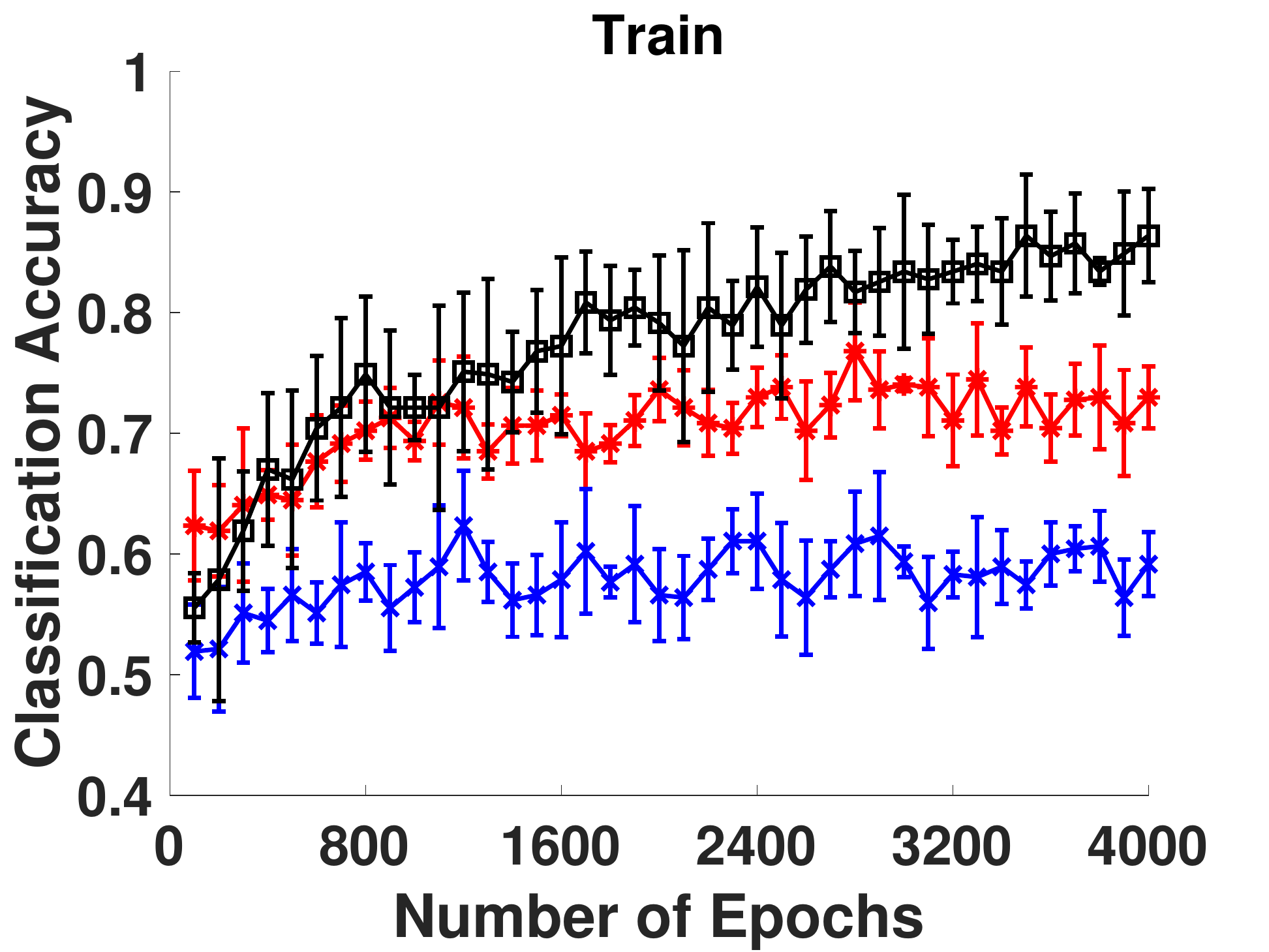}
        \includegraphics[width=0.24\linewidth,keepaspectratio]{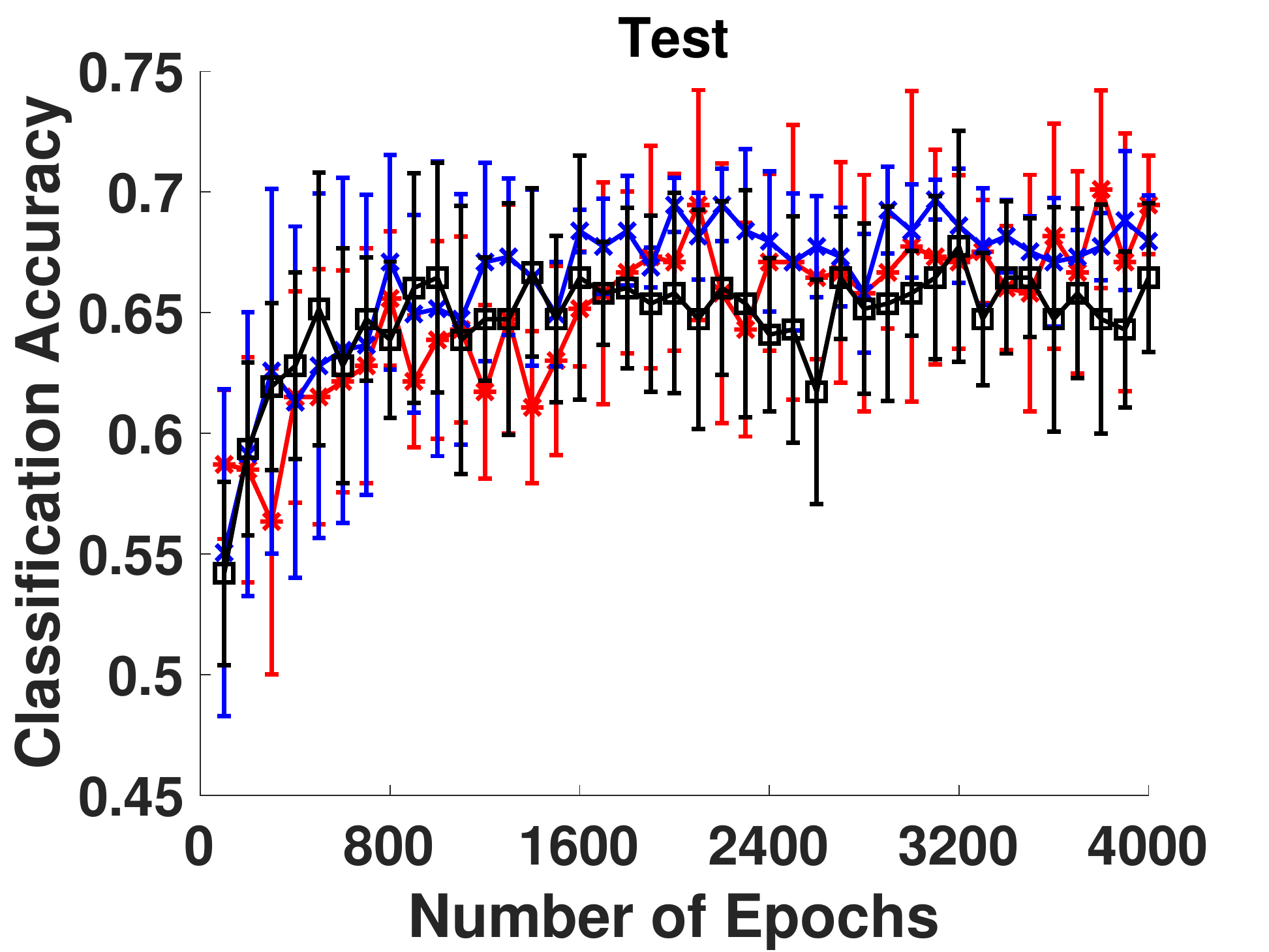}\\
        (E) Prostate\_GE & (F) SMK\_CAN
        \end{tabular}
        \caption{Comparison among FsNet, supervised CAE, Diet Network, and SVM in terms of mean training and testing accuracies over the epochs. For the neural-network-based approaches, we set the model parameters to $b = 10$ and $K = 10$. }
        \label{fig:allacc_full}
        \vspace{-.15in}
    \end{figure*}

    \begin{figure*}[!ht]
        \centering
        \begin{tabular}{c c c}
            \includegraphics[width=0.3\linewidth,keepaspectratio]{figs/ALLAML_test_recon.pdf}&
            \includegraphics[width=0.3\linewidth,keepaspectratio]{figs/CLL_SUB_111_test_recon.pdf}&
            \includegraphics[width=0.3\linewidth,keepaspectratio]{figs/GLI_85_test_recon.pdf}\\
            (A) ALLAML & (B) CLL\_SUB & (C) GLI\_85\\
            \includegraphics[width=0.3\linewidth,keepaspectratio]{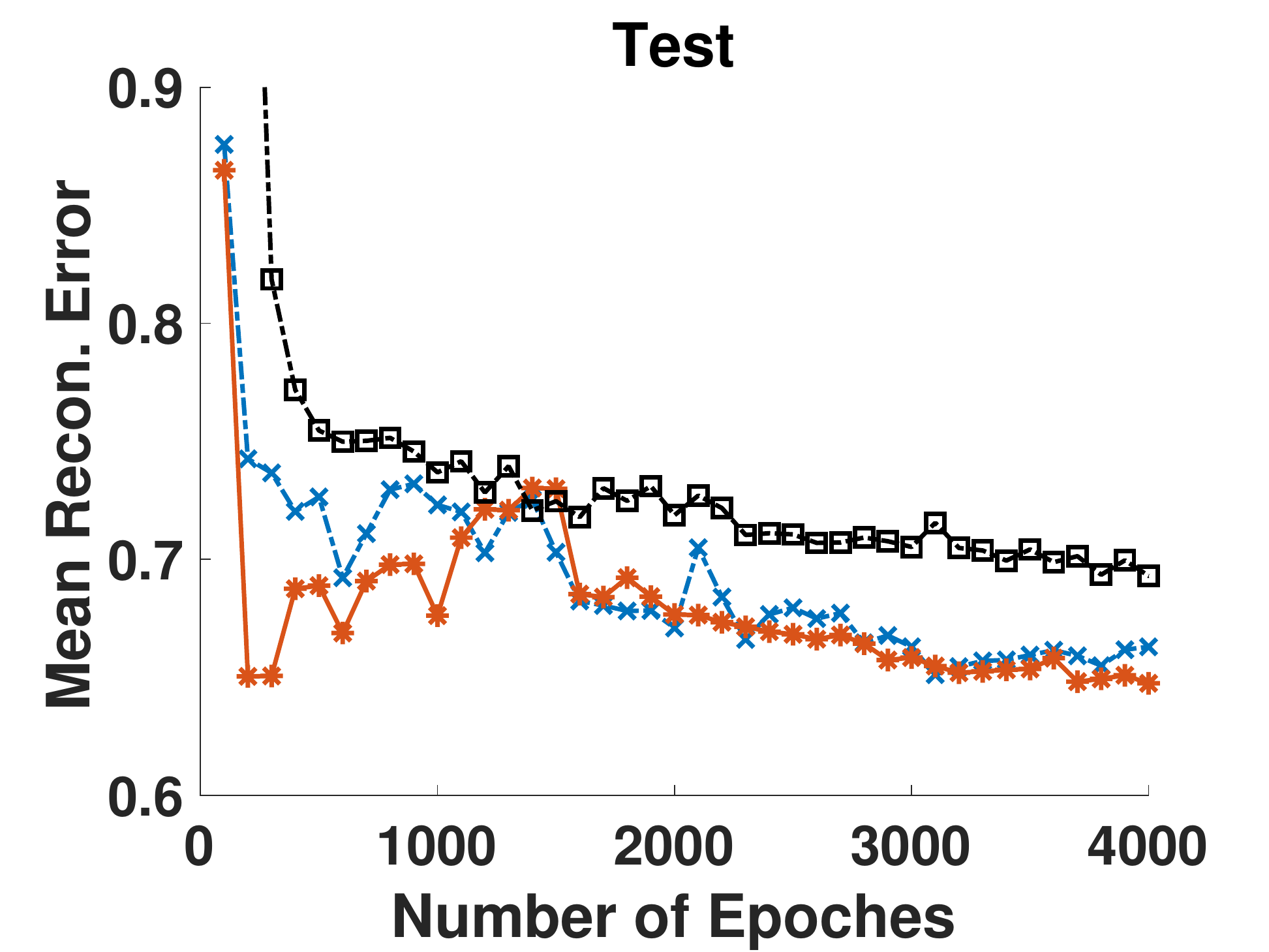}&
            \includegraphics[width=0.3\linewidth,keepaspectratio]{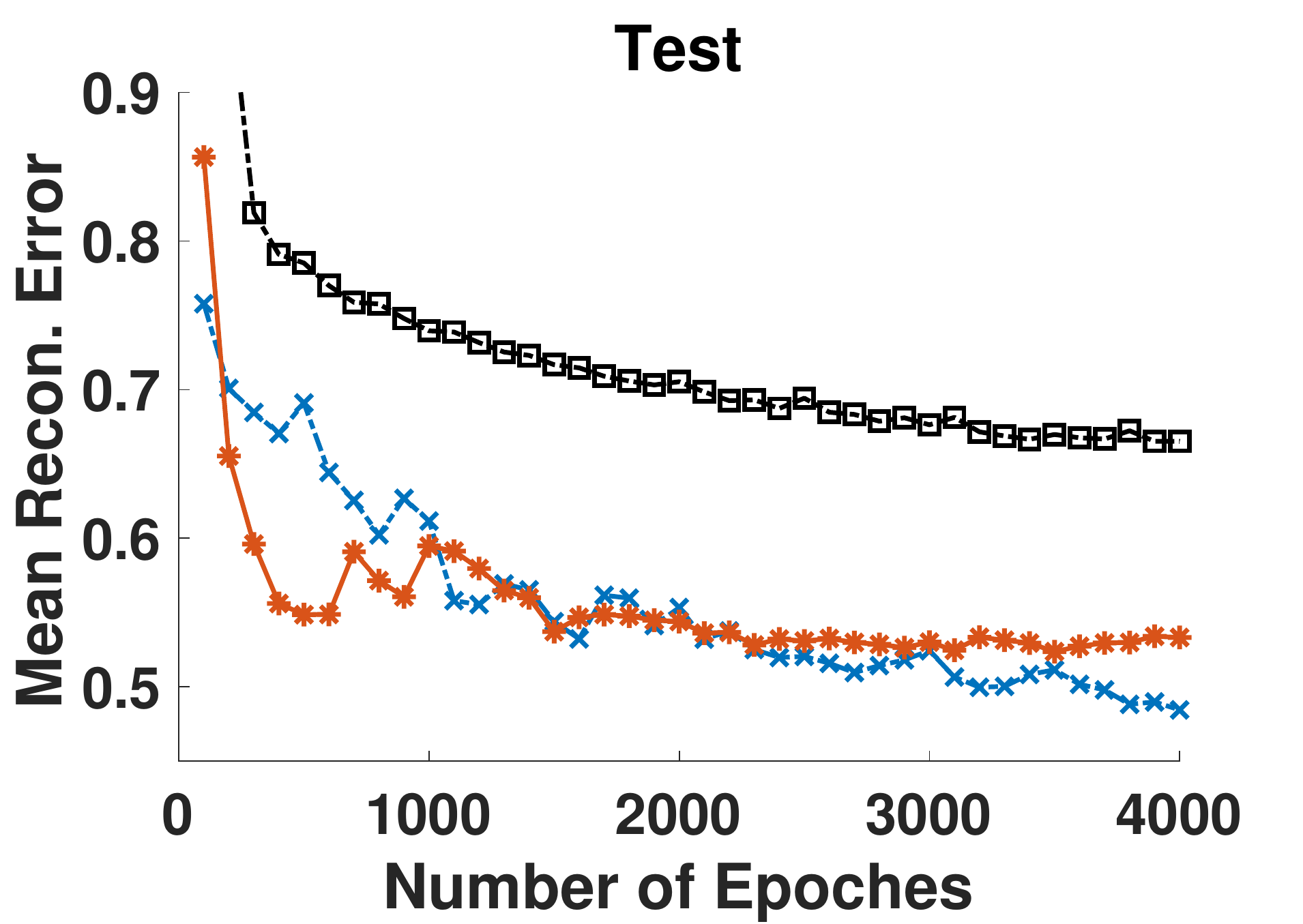}&
            \includegraphics[width=0.3\linewidth,keepaspectratio]{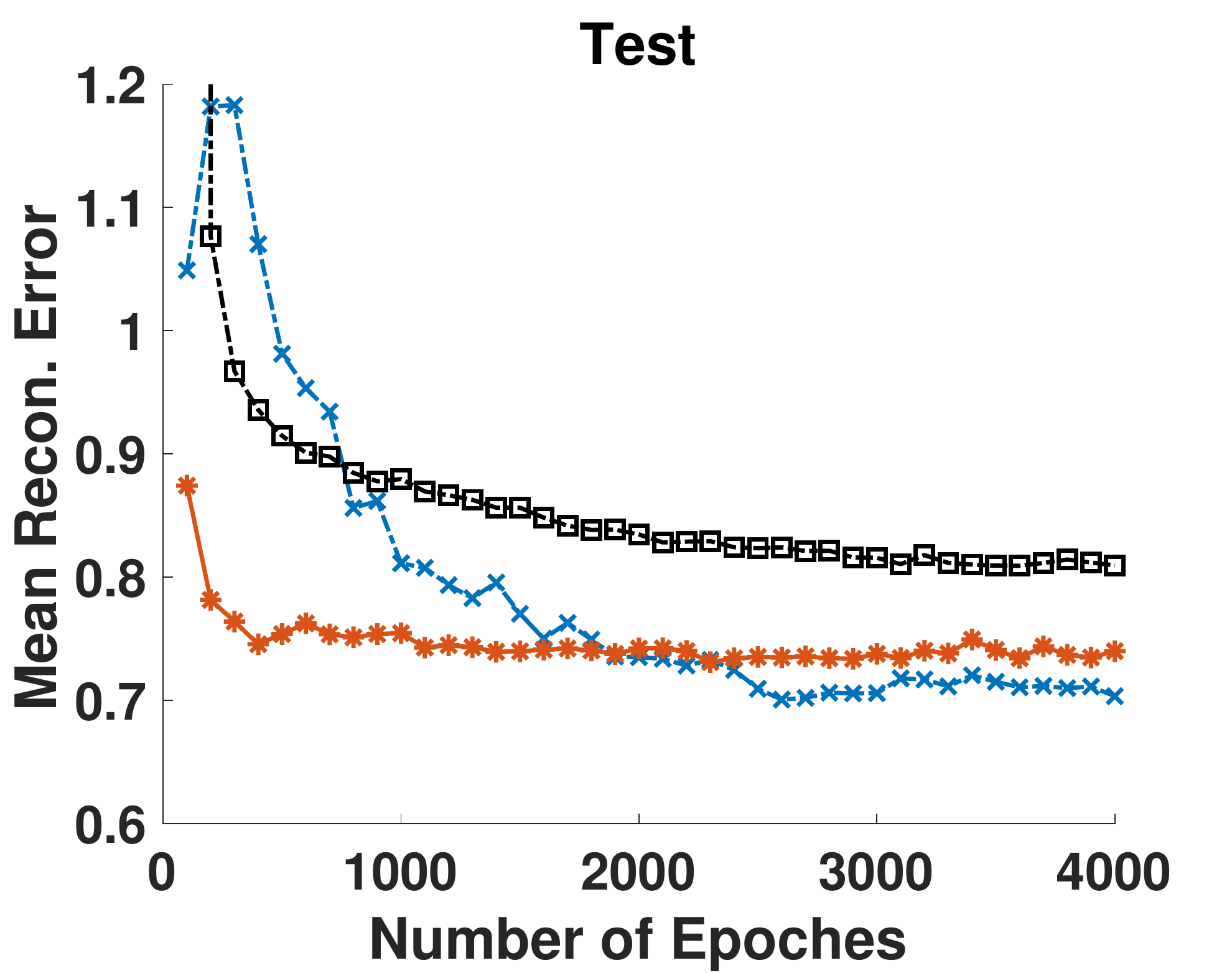}\\
            (D) GLIOMA & (E) Prostate\_GE & SMK\_CAN
        \end{tabular}
        \caption{Comparison between the proposed FsNet and existing supervised CAE approaches in terms of the mean test reconstruction error over the epochs.}
        \label{fig:allrecon_all}
    \end{figure*}

\end{document}